\documentclass{article} % For LaTeX2e
\usepackage{iclr2025_conference,times}

\usepackage{amsmath,amsfonts,bm}

\def\eqref#1{equation~\ref{#1}}
\def\1{\bm{1}}

\DeclareMathAlphabet{\mathsfit}{\encodingdefault}{\sfdefault}{m}{sl}
\SetMathAlphabet{\mathsfit}{bold}{\encodingdefault}{\sfdefault}{bx}{n}

\usepackage{hyperref}
\usepackage{url}

\usepackage[utf8]{inputenc} % allow utf-8 input
\usepackage[T1]{fontenc}    % use 8-bit T1 fonts
\usepackage{hyperref}       % hyperlinks
\usepackage{url}            % simple URL typesetting
\usepackage{booktabs}       % professional-quality tables
\usepackage{amsfonts}       % blackboard math symbols
\usepackage{nicefrac}       % compact symbols for 1/2, etc.
\usepackage{microtype}      % microtypography
\usepackage{xcolor}         % colors

\usepackage[utf8]{inputenc} % allow utf-8 input
\usepackage[T1]{fontenc}    % use 8-bit T1 fonts
\usepackage{url}            % simple URL typesetting
\usepackage{booktabs}       % professional-quality tables
\usepackage{amsfonts}       % blackboard math symbols
\usepackage{nicefrac}       % compact symbols for 1/2, etc.
\usepackage{microtype}      % microtypography
\usepackage{xcolor}         % colors

\usepackage{amsmath}
\usepackage{amssymb}
\usepackage{wrapfig}
\usepackage{subfig}
\usepackage{caption}
\usepackage{color}
\usepackage{algorithm}
\usepackage{algorithmic}
\usepackage{graphicx}

\usepackage{pgfplots}
\usepackage{tikz}
\usepackage{tikz-3dplot}

\usepackage{ragged2e}
\usepackage{fp}

\usepackage{ifthen}

\usepackage{tikz}
\usetikzlibrary{shapes.geometric, arrows, calc,automata,positioning,decorations,decorations.text}

\usetikzlibrary{mindmap,snakes}

\usepackage{multirow}

\usepackage{algorithm}
\usepackage{algorithmic}
\usepackage{graphicx}
\usepackage[english]{babel}
\usepackage{array}
\usepackage[export]{adjustbox}

\usepackage{nicefrac}

\usepackage{xcolor}

\usepackage{colortbl}
\usepackage{makecell}
\usepackage{pifont}

\usepackage{booktabs}
\newcommand{\cmark}{\ding{51}}%
\newcommand{\xmark}{\ding{55}}

\graphicspath{{images/}}

\usepackage[font=small]{caption}

\usepackage{mathptmx}
\usepackage{bm}

\usepackage[outline]{contour}

\definecolor{mycitecolor}{rgb}{0.005, 0.3, 0.7}
\usepackage{url}

\colorlet{accclr}{white!20!black}
\colorlet{flopclr}{green!40!blue}
\colorlet{flopclr}{black!10!flopclr}

\colorlet{baseclr}{gray}
\colorlet{rwclr}{white!90!baseclr}
\colorlet{cellclr}{white!90!baseclr}

  \colorlet{dotaclr}{white}
\colorlet{dotbclr}{magenta}

\newcommand{\bdota}{\textcolor{dotaclr}{$\bullet$}}
\newcommand{\bdotb}{\textcolor{dotbclr}{$\bullet$}}

\colorlet{xmarkclr}{cyan}
\colorlet{cmarkclr}{magenta}

\newcommand{\buac}[1]{\textcolor{#1}{\raisebox{0.22ex}{\protect\contour{#1}{$\uparrow$}}}}  % bold math symbol
\newcommand{\bdac}[1]{\textcolor{#1}{\raisebox{0.22ex}{\protect\contour{#1}{$\downarrow$}}}}  % bold math symbol

\definecolor{mylinkcolor}{rgb}{0.005, 0.3, 0.7}
\definecolor{suppcolor}{rgb}{0.005, 0.3, 0.5}

\definecolor{mylinkcolor}{rgb}{1.0, 0.2, 1.0}
\newcommand{\OursModule}{OSMA}
\newcommand{\OursModuleFull}{One Step Multiscale Attention}
\newcommand{\OursModuleb}{CRAM}
\newcommand{\OursModuleFullb}{Cross Resolution Attention Module}
\newcommand{\Ours}{CRED}
\newcommand{\OursFull}{Cross-Resolution Encoding-Decoding}

\newcommand{\halfres}{DC$\times0.25$}

\title{Cross Resolution Encoding-Decoding For\\ Detection Transformers}

\author{Ashish Kumar \\
General Autonomy\\
Bangalore, India \\
\texttt{ashish@generalautonomy.in} \\
\And
Jaesik Park \\
Seoul National University \\
Seoul, South Korea \\
\texttt{jaesik.park@snu.ac.kr} \\
}

\iclrfinalcopy % Uncomment for camera-ready version, but NOT for submission.

\begin{document}

\maketitle

\vspace{-2ex}

\begin{abstract}
\vspace{-0.5ex}
Detection Transformers (DETR) are renowned object detection pipelines, however computationally efficient multiscale detection using DETR is still challenging. In this paper, we propose a \OursFull{} (\Ours{}) mechanism that allows DETR to achieve the accuracy of high-resolution detection while having the speed of low-resolution detection. \Ours{} is based on two modules; \OursModuleFullb{} (\OursModuleb{}) and \OursModuleFull{} (\OursModule{}). \OursModuleb{} is designed to transfer the knowledge of low-resolution encoder output to a high-resolution feature. While \OursModule{} is designed to fuse multiscale features in a single step and produce a feature map of a desired resolution enriched with multiscale information. When used in prominent DETR methods, \Ours{} delivers accuracy similar to the high-resolution DETR counterpart in roughly $50\%$ fewer FLOPs. Specifically, state-of-the-art DN-DETR, when used with \Ours{} (calling \Ours{}-DETR), becomes $76\%$ faster, with $\sim50\%$ reduced FLOPs than its high-resolution counterpart with $202$G FLOPs on MS-COCO benchmark. We plan to release pretrained \Ours{}-DETRs for use by the community. Code: \textcolor{mylinkcolor}{\url{https://github.com/ashishkumar822/CRED-DETR}}.

\end{abstract}

\vspace{-1ex}

\section{Introduction}
Detection Transformers (DETR) \cite{detr} are end-to-end object detection frameworks without post-processing, such as anchor boxes, box matching, and non-maximal suppression (NMS) that are required in ConvNet-based detectors \cite{fasterrcnn, ssd}. Since their first entry \cite{detr}, DETRs have evolved, mainly regarding query design~\cite{smcdetr, deformabledetr, conddetr, anchordetr, dabdetr, dndetr} to improve the detection performance, while recent DETR pipelines achieve that via salient points \cite{sapdetr} or unsupervised pretraining \cite{siamesedetr}.

State-of-the-art DETRs exploit high-resolution features (known as Dilated Convolution or DC variant) \cite{dabdetr, dndetr} or dense multi-scale features \cite{dinodetr} to push the detection accuracy further. However, they suffer from high computation complexity w.r.t. their accuracy gains. For example, DAB DETR~\cite{dabdetr} with high-resolution features improved $2.3$AP, but it introduces a $114\%$ rise in FLOPs. 

The primary reason is the quadratic computational complexity \cite{coatnet} of the Transformer's attention mechanism \cite{imfa}  w.r.t. the spatial size, i.e., a Transformer is $O (H^2 W^2)$ complex in processing a feature map of a spatial size of $H \times W$. Hence, doubling the resolution of the encoder input quadruples its computations while also affecting the decoder. Although deformable attention \cite{deformabledetr} addresses this issue, it causes additional runtime overhead due to irregular memory accesses.

\begin{figure}[t]
\includegraphics[scale=0.313]{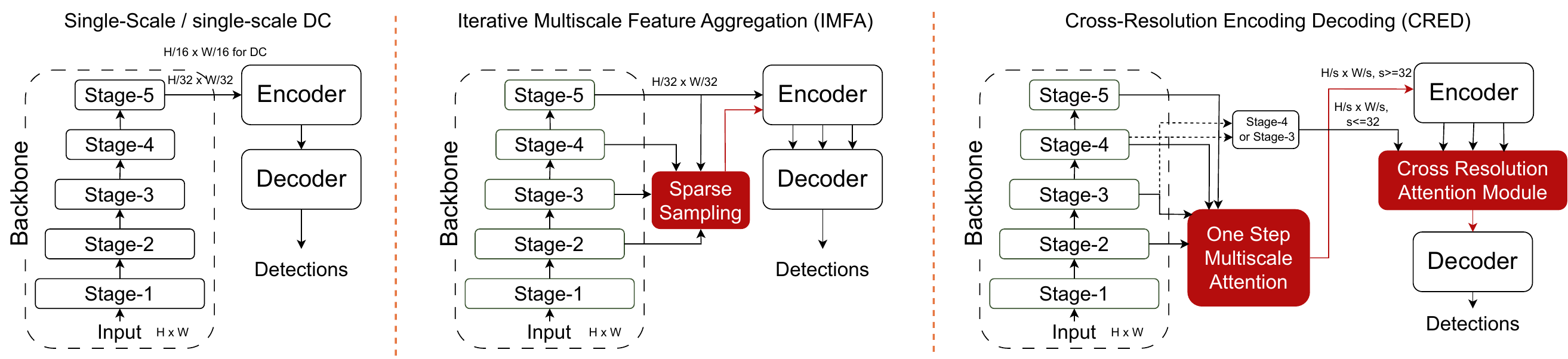}
\caption{Left: Single-Scale and/ DC DETR. Middle: IMFA DETR \cite{imfa}. Right: CRED DETR (Ours). Multiple arrows between two modules indicate layerwise refinement. Stage-$1$ features are generally not used due to large resolution and small receptive field.}
\label{fig:cred_in_detr}
\vspace{-2ex}
\end{figure}

To address this issue, recent IMFA \cite{imfa} proposes using a low-resolution feature map and Top-K sparsely sampled higher-resolution features (Figure~\ref{fig:cred_in_detr}). IMFA exhibits improvements in accuracy with lower FLOPs. However, sparse sampling incurs memory access costs due to irregular memory access, which becomes prevalent with more samples. 
Recent \cite{litedetr} exploits attention among only interleaved tokens for reducing computations in multiscale attention in the encoder. \cite{detrbeatsyolo} works on reduced resolution images ($640 \times 640$) instead of high resolution settings ($800 \times 1200$) on standard DETR.

Our aim aligns with improving DETR speed and accuracy by using multiscale features at high-resolution settings. However, our key motivation is based on our finding that the encoder consists of most computations relative to the decoder. Therefore, we propose a principal design change in the DETR pipeline, i.e., feeding the encoder with low-resolution while feeding the decoder with a high-resolution feature obtained from the backbone. By keeping the encoder input low-resolution, we save computations, while by keeping the decoder input high-resolution, we provide the decoder access to fine-grained details. Since the high-resolution features from the backbone lack large spatial context~\cite{detr}, we develop a novel \OursModuleFullb{} (\OursModuleb{}). It utilizes encoder output that has a global context and transfers this information into the high-resolution feature map. When this feature map is fed to the decoder, the decoder has access to the fine-grained details and the global context, thus improving the accuracy and runtime.

Then, by exploiting this capability of \OursModuleb{} to transfer the information from low-resolution encoder output to high-resolution feature, we propose to reduce the resolution of an encoder further to save computations. This behavior is intended to develop faster DETRs offering speed-accuracy tradeoffs. However, feeding the encoder naively with reduced resolution degrades its performance. Hence, we devise a novel module called \OursModuleFull{} (\OursModule{}), which attends to multiscale information in one step and can produce tokens or feature maps of the desired resolution enriched with multiscale information. When the encoder is fed the tokens produced by \OursModule{} at aggressively low resolution, accuracy is considerably improved while avoiding any runtime overhead relative to the baseline which was fed with a low-resolution backbone output.

We name our overall approach as \OursFull{} (\Ours{}), shown in Figure~\ref{fig:cred_in_detr}. We demonstrate that our \Ours{}-enhanced DETR can attain an AP (average precision) equivalent to the original DETR's high-resolution counterpart in $50\%$ fewer FLOPs at $76\%$ improved runtime. For instance, DN-DETR \cite{dndetr}+\Ours{} reduces FLOPs from $202$G to $103$G ($\sim50\%$\bdac{black}) and improves FPS from $13$FPS to $23$FPS ($\sim76\%$\buac{black}) without losing accuracy. In addition, applying \Ours{} in DETR variants \cite{conddetr, dabdetr, imfa} consistently improves their accuracy and runtime compared with their DC variants. 
To improve runtime further, we half the encoder resolution via \OursModule{}. Interestingly, we only observe $-1$AP in this configuration while the runtime is further improved by $84\%$ compared with the vanilla DC variant. This signifies the potential of \OursFull{} mechanism in DETRs.

%%%%
%%%%
%%%%
\section{Preliminary}
\label{sec:detr_primer}
This section revisits the architectural design of vanilla DETR~\cite{detr} and advanced DETRs~\cite{dabdetr, dndetr, deformabledetr}. 
DETR comprises a backbone, a Transformer encoder, and a Transformer decoder. 
In a backbone \cite{dndetr}, it is common to keep five stages, each operating at a resolution half of its previous stage. 
Thus, the final stage (stage-$5$) runs at a \textit{stride of 32} at the original resolution.
Once an image $I\in \mathbb{R}^{3 \times H \times W}$ is fed into a backbone (Figure~\ref{fig:cred_in_detr}), the backbone output $F_b \in \mathbb{R}^{C \times H_0 \times W_0}$ is fed to a Transformer encoder, producing encoded feature embeddings or tokens $ F_e \in \mathbb{R}^{d \times H_0 \times W_0}$. These embeddings are fed to the Transformer decoder to produce a fixed number of queries ($N_q$), each representing an object detectable in the image. The queries are passed through two Feedforward neural Networks (FFN) to obtain the object class and its bounding box. During training, bipartite or Hungarian matching \cite{detr} is performed for one-to-one assignments of ground truth and predictions.

In DETR pipelines \cite{detr, dabdetr, dndetr}, the embeddings $F_e$ are directly fed to the decoder where the $N_q$ queries interact with $H_0\times W_0$ features in $F_e$ via cross attention. The encoder complexity is $O(H^2_0W^2_0)$ in its self-attention, whereas for the decoder, it is $O(N_q H_0W_0)$ in cross attention and $O(N^2_q)$ in its self-attention. In DETRs, $N_q$ is relatively smaller, i.e., $300$; however, $H_0W_0$ is quite a large number depending on the image size, e.g., $1280 \times 800$ \cite{dabdetr, dndetr}. Hence, \textit{most of the computations are concentrated in the encoder}. When opting for high-resolution detection, known as the DC variant \cite{dabdetr,dndetr}, the backbone output stride is set lower than $32$. For this purpose, stage-$5$ of the backbone is set to run at \textit{stride 16} \cite{conddetr, dabdetr, dndetr} w.r.t. the image, thereby doubling the resolution of $F_b$. This leads to quadratic growth in the encoder computations due to the increased resolution \cite{coatnet}.  

In this paper, we rethink the encoder-decoder information flow while leveraging multiscale features in a computationally efficient manner. Summarily, we propose an approach that feeds the encoder with a stride $\geq 32$ while feeding the decoder with a stride $\leq 32$ to harness the best of both worlds, i.e., the accuracy of high-resolution detection and speed of low-resolution detection. To our knowledge, such a mechanism has not been demonstrated in DETRs.
% , which is our main contribution.
% The exact details are discussed in the next section.

%%%%
%%%%
%%%%
\section{Method}
\label{sec:method}
\label{sec:method}
We propose to enhance the DETR design with the two modules. \textit{Firstly}, we develop \OursModuleb{} which enables \OursFull{}. \textit{Secondly}, we develop \OursModule{} module, which facilitates generating feature maps enriched with multiscale information in a computationally efficient manner and further enhances the performance of \Ours{}. 

\subsection{Cross Resolution Attention Module (\OursModuleb{})}
\label{sec:cram}
\begin{figure}[!t]
\centering
\includegraphics[scale=0.47]{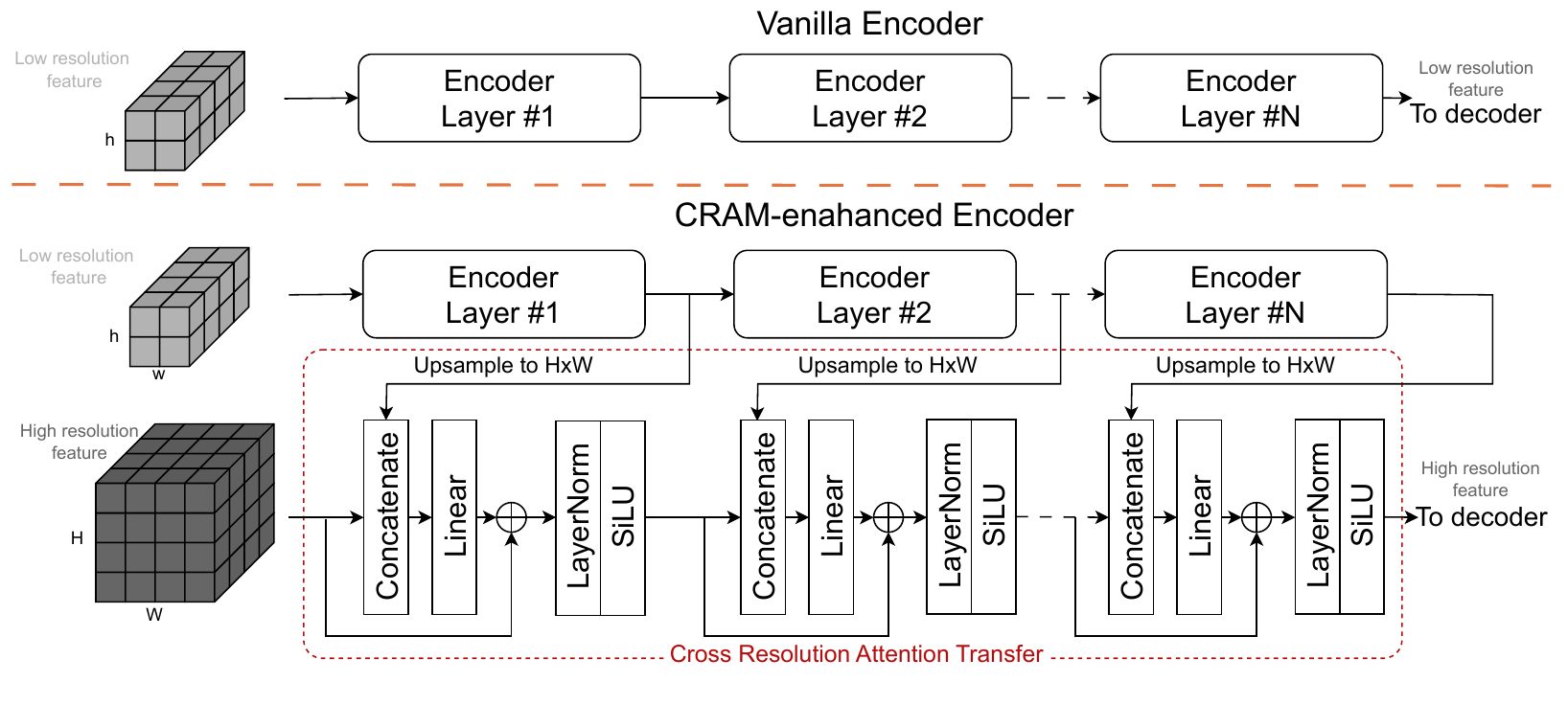}
\vspace{-1ex}
\caption{\OursModuleb{}: Cross Resolution Attention Module.}
\label{fig:cram}
\vspace{-2ex}
\end{figure}

As mentioned in Sec.~\ref{sec:detr_primer}, high-resolution detection improves DETR accuracy, but high-resolution input to the encoder is compute-intensive due to its quadratic complexity \cite{coatnet}. Hence, we feed the encoder with low resolution while feeding the decoder with high resolution, calling it \textit{cross-resolution encoding-decoding paradigm}. However, the information flow between them no longer exists due to the different input sources to the encoder and decoder. 
Hence, we develop \OursModuleb{} that acts as a bridge between the encoder and decoder in the proposed cross-resolution encoding decoding paradigm.
The overall design of our approach (\OursModuleb{}) is shown in Figure~\ref{fig:cram}.

Consider two feature maps $X \in \mathbb{R}^{C \times h \times w}$ and $Y \in \mathbb{R}^{C \times H \times W}$. In our \OursFull{} approach, the low-resolution feature $X$ is fed to the encoder layers, while the high-resolution feature $Y$ is fed to the attention transfer module. To transfer the knowledge of the encoder embeddings ($X_e = \text{Encoder}(X),~X_e \in \mathbb{R}^{C \times h \times w}$) to $Y$, we spatially upsample ($\hat{X}_e = \text{Upsample}(X_e) |_{H, W} ~\hat{X}_e \in \mathbb{R}^{C \times H \times W}$) the encoder output to match the resolution of $Y$, and a concatenation operation is performed.
% Mathematically, we write the operation as below:
%
% \begin{align}
% X_e &= \text{Encoder}(X),~~~X_e \in \mathbb{R}^{C \times h \times w} \\
% \hat{X}_e &= \text{Upsample}(X_e) |_{H, W} ~~~\hat{X}_e \in \mathbb{R}^{C \times H \times W}
% \end{align}
%

Now, we use a linear projection layer, which combines the features information of the high-resolution features and the upsampled low-resolution feature from the encoder. The output ($Z = \text{Linear}(\text{Cat}(Y, X_e))$) is normalized via LayerNorm \cite{layernorm} and passed through SiLU activation \cite{silu} ($\hat{Y}_e = \text{SiLU} (\text{LayerNorm}(Z)), ~\hat{Y}_e \in \mathbb{R}^{C \times H \times W}$). We use the residual connections to facilitate smoother optimization. %These operations can be summarized as below:
%
% \begin{align}
% Z &=  \text{Conv}_{1\times1}(\text{Cat}(Y, X_e)),~~~~
%  \hat{Y}_e = \text{SiLU} (\text{LayerNorm}(Z)), ~~~ \hat{Y}_e \in \mathbb{R}^{C \times H \times W}
% \end{align}
%
This process is repeated as many times as the number of encoder layers to refine the high-resolution $Y$ by transferring the knowledge embedded in the encoder layers. The input to \OursModuleb{} is initialized with the high-resolution feature from an intermediate backbone stage (Sec.~\ref{sec:detr_primer}), whereas the final $\hat{Y}_e$ is fed to the decoder for the predictions

\paragraph{Computation Complexity}
An encoder layer has $\mathcal{O}(h^2w^2)$ complexity to process its input $X_e \in \mathbb{R}^{h \times w}$. In our case, we feed the encoder with 
$X_e \in \mathbb{R}^{\nicefrac{H}{r} \times \nicefrac{W}{r}}$, where $r \ge 2$, and feed the decoder with $X_d \in \mathbb{R}^{H \times W}$. In aggregation, we have a total complexity of $\mathcal{O}(\frac{H^2W^2}{r^4} + HW)$, which is far lower than the vanilla encoder complexity of $\mathcal{O}(H^2W^2)$. With this strategy, we achieve accuracy equivalent to when the encoder is fed with the high-resolution feature while having at least $50\%$ fewer overall computations for $r=2$. Specifically, w.r.t. vanilla encoder with high resolution, a FLOP saving of $50\%$ is obtained. See Sec~\ref{sec:ablations} for the computational complexity analysis.

\paragraph{Why Could Cross Resolution Attention Transfer Improve Performance?} In vanilla DETR design, the encoder embeddings are produced from stage-$5$ (low-resolution) and have global receptive filed via self-attention \cite{vit}, which are fed to the decoder for the predictions. However, these embeddings do not have fine details of smaller objects. On the other hand, the high-resolution input $Y$ (earlier stages, e.g., stage-$4$) to \OursModuleb{} has a small receptive field but has details of smaller objects.

In \OursModuleb{}, the concatenation operation followed by a linear layer infuses the local and global context to produce fine-grained, high-resolution features, similar to what earlier semantic segmentation approaches \cite{pspnet} used for improving performance. In the same way, with this operation, the high-resolution feature $Y$ acquires a global receptive field when concatenated with the encoder embedding or tokens $X_e$. After the layerwise refinement, it is fed to the decoder, which improves the accuracy and speed; even the encoder still functions at a smaller resolution.

\begin{figure}
\centering 

\includegraphics[scale=0.325]{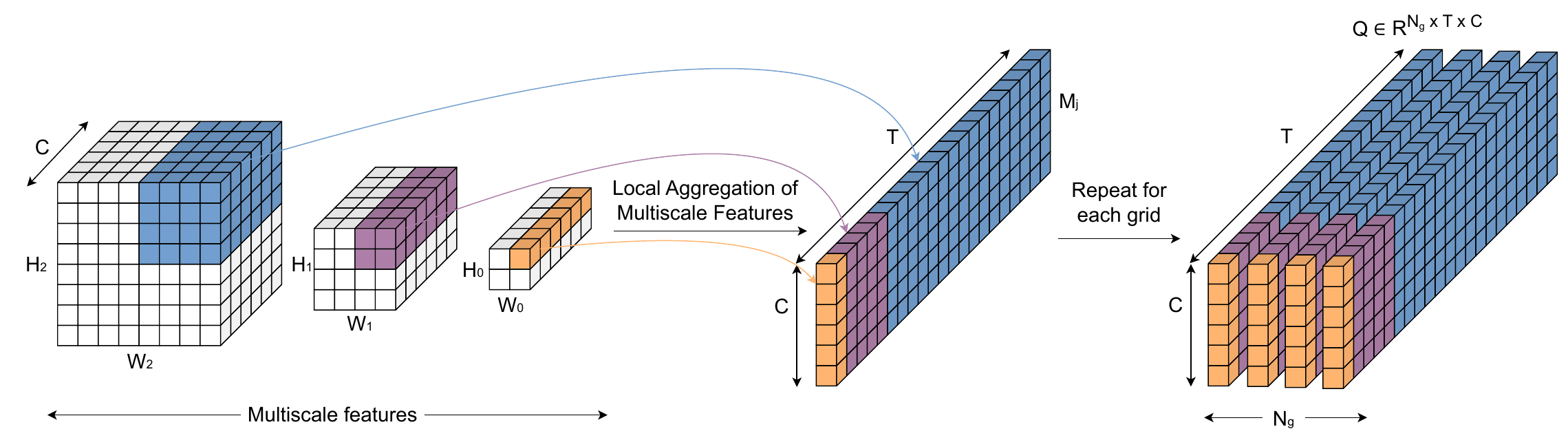}

\caption{Local aggregation of the multiscale features in \OursModule{} for $g_0=1$ which produces $Q \in \mathbb{R}^{N_g \times T \times C}$}

\label{fig:osma_step1}

\vspace{-2ex}
\end{figure}
\subsection{\OursModuleFull{} (\OursModule{})}
\label{sec:osma}
In single-scale operation of DETRs \cite{dabdetr, dndetr}, the backbone output $F_b$ is fed to the encoder. However, $F_b$ does not have direct access to the multiscale information. Whereas multiscale DETRs feed the encoder with either sampled \cite{deformabledetr} or dense multiscale features \cite{dinodetr}, which is computationally heavy.

We propose \OursModule{}, which produces $F_o$ enriched with multiscale information, i.e., the best of both single-scale and multiscale methods without progressive fusion, i.e., fusing two scales at a time~\cite{yolov8}(Figure~\ref{fig:cred_in_detr}). In general, $F_b$ has a stride $32$ w.r.t. the input \cite{resnet}; however, using multiscale features, \OursModule{} can produce a feature map $F_o$ (`o' refers to \OursModule{}) of stride greater or lesser than $32$ which directly controls the encoder computations. With this functionality, \OursModule{} offers better features or tokens enriched with multiscale information to be utilized by the encoder.

\OursModule{} has three main steps: \textit{First}, Local aggregation of multiscale features, \textit{Second}, performing one-step attention on the aggregated features, and \textit{Third}, broadcasting the output into desired resolution.

\noindent\textbf{Local Aggregation of Multiscale Features.}
This step aggregates $n$ multiscale features $F^i \in \mathbb{R}^{C\times H_i \times W_i}$ obtained from the backbone, 
where $i=0,1,...n-1$ is the scale index (Figure~\ref{fig:osma_step1}).
% , `$0$' denoting the lowest resolution 
In this step, all the features are first divided into non-overlapping grids of size $g_i=2^i*g$. The total number of grids for all the scales is equal and is given by $N_g = \nicefrac{H_i}{g_i} \times \nicefrac{W_i}{g_i}$.

Let $F^i_{j} \in \mathbb{R}^{C\times g \times g}$ denotes feature map of $j^{th}$-grid of $F^i$, where $j=\{0,1,... N_g\}$. For each scale, we flatten $F^i_{j}$ in spatial dimension and stack all its features with its corresponding grids in other scales, as shown in colors in Figure~\ref{fig:osma_step1}. This results in a matrix $M_j \in \mathbb{R}^{C \times T}$ for each grid. 
% Mathematically, if $F^i_{jk} \in \mathbb{R}^{C}$ denotes $k^{th}$ feature in the map $F^i_j$, the aggregation operation can be written as below:
%
% \begin{align}
%  M_j = [F^0_{j0},...,F^0_{jk_0-1}, F^1_{j0},...,F^1_{jk_1-1},...,F^{n-1}_{j0},...,F^{n-1}_{jk_{n-1}-1}],~~~~~~ k_i = g_i \times g_i
%  \label{eq:matrixM}
% \end{align}
%
% where, $M_j \in \mathbb{R}^{C \times T}$
% , and $T=\sum\limits^{i\in n}k_i$. 
This step is repeated for each non-overlapping grid, and we obtain $N_g$ matrices, referring as $Q \in \mathbb{R}^{N_g \times T \times C}$. The number $N_g$ is absorbed into the batch dimension while processing batched training or inference so that all the matrices can be processed in parallel. 
This step requires all the feature maps $F^i$ to be an integral multiple of $g$. Hence, we align all the feature maps w.r.t. $g_i$ before performing local aggregation via bilinear interpolation. See Table~\ref{tab:ablation_osma} for ablation on `$g$'.

\begin{figure}[t]
\centering
\includegraphics[scale=0.25]{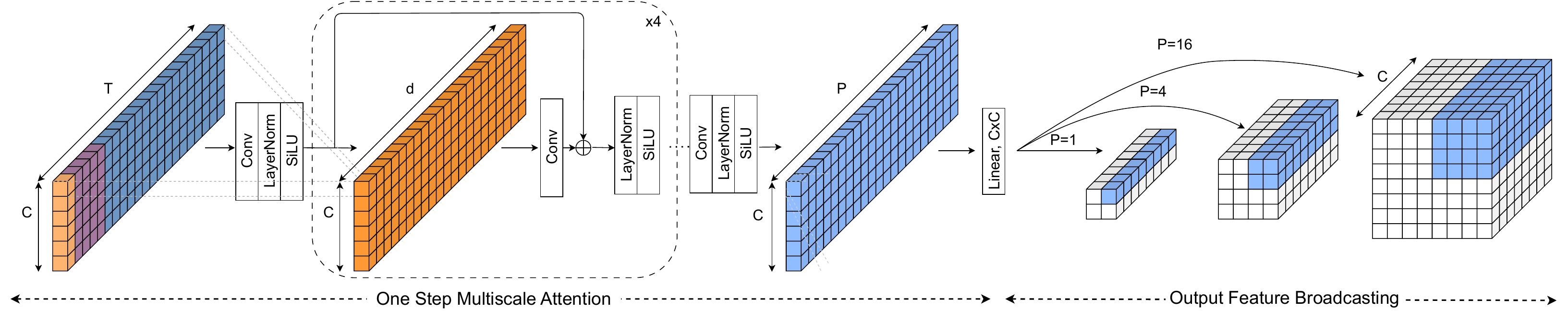}

\caption{One step attention and Output Broadcasting in \OursModule{}. `One-Step' should not be confused with `single layer'; instead, it refers to all the multiscale features being attended simultaneously through $1\times 1$ layer. Output broadcasting infers the shape of the output feature based on the value of $P$.}
\vspace{-2ex}
\label{fig:osma_step_2_3}
\end{figure}

% \subsubsection{One Step Multiscale Attention}
\noindent\textbf{One Step Multiscale Attention.}
We perform information fusion by attending to all the multiscale features and their channels simultaneously, thus calling it a one-step multiscale attention. We describe this process for a single grid or $M_j$, also depicted in Figure~\ref{fig:osma_step_2_3}.
\par
In the attention process, $M_j$ undergoes a $1\times 1 $ convolution operation with weights $\mathbf{W} \in \mathbb{R}^{d \times T \times 1 \times 1}$ which projects $M_j$ in a $d$ dimensional latent space. The $1\times 1$ layer combines information from all the $T$ multiscale features with a unit stride. In this way, the output of multiscale information becomes richer. Then, we perform a layer normalization over the columns of $M_j$ followed by SiLU activation (See Sec.~\ref{sec:ablations}). 
% Mathematically, we can summarize the process as below:
% %
% \begin{align}
% \hat{M}_j = \text{SiLU}(\text{LayerNorm}(\text{Conv}(M_j;\mathbf{W},\mathbf{b})) 
% \end{align}
%

This process is repeated for feature refinement, and the penultimate $1\times 1$ layer produces $P$ channels. The final layer is a linear projection applied over the columns of $\hat{M}_j$ to refine the column information because a column becomes a feature $\in \mathbb{R}^C$ in the output feature map $F_o$. The output of this step $\hat{Q} \in \mathbb{R}^{N_g \times P \times C}$ is broadcasted into a feature based on the requirements, as discussed next. 
%
%
% \subsubsection{Output Feature Broadcasting}
\noindent\textbf{Output Feature Broadcasting.}
\label{sec:inverseattention}
This step broadcasts $\hat{Q}$ into a feature map $F_o$ of the desired resolution based on the pair $\{g_0,P\}$. For example, for $g_0=1$, if we aim to produce a feature $F_o$ of size $H_0 \times W_0$, the value of $P$ will be set to $1$, or if a feature $F_o$ of size $2H_0 \times 2W_0$ is required, $P$ can be set to $4$, or if a feature $F_o$ of size $H_0/2 \times W_0/2$ is needed, $\{g_0=2, P=1\}$ can be used. %Similarly, other resolutions can be constructed. 

This flexibility of generating $F_o$ of desired resolution allows controlling the encoder's input resolution and, hence, its computations. Meanwhile, multiscale information infusion helps improve DETR accuracy with a slight computational overhead. We have studied various $\{g_0,P\}$ combinations, and our empirical results show that $\{g_0=1,P=1\}$ is the best combination for keeping the resolution of $F_o$ equals to $F_0$ whereas $\{g_0=2, P=1\}$ is best for reducing the resolution. See Table~\ref{tab:ablation_osma}.

\subsection{Configuring \Ours{} for DETRs}

\label{sec:cred_config}

We mainly test two important configurations. In all the configurations, \OursModule{} feeds the encoder, whereas \OursModuleb{} feeds the decoder. \OursModule{} is always fed with $\{F^3,F^4,F^5\}$ output of backbone (Figure~\ref{fig:cred_in_detr}). Each configuration has different settings for \OursModule{} and input sources for \OursModuleb{}.

\textbf{Default.} \OursModule{} operates at $\{g_0=1, P=1\}$ i.e. its output resolution is same as $F^5$. Whereas \OursModuleb{} is fed with $F^4$. In other words, the encoder runs at half the resolution of the decoder.

\textbf{Configuration: `\halfres{}'.} \OursModule{} operates at $\{g_0=2, P=1\}$, producing feature map of half the resolution of $F^5$ or quarter of $F^4$ or quarter of DC. Whereas \OursModuleb{} is fed with $F^4$. This configuration evaluates the capability of our \OursFull{} when the encoder is fed with a very low-resolution map. For comparison, the encoder in baselines is fed with $F^5$ downsampled to half of its resolution. Since this resolution is $\nicefrac{1}{4}$ of DC resolution, it is named as \halfres{}.

\textbf{Additional Configuration: `OO'} This is similar to the default configuration except the input source to \OursModuleb{}. We use two instances of \OursModule{}: \OursModule{}$_e$ and \OursModule{}$_c$. The former operates at $\{g_0=1, P=1\}$, feeding the encoder, while the latter operates at $\{g_0=1, P=4\}$, i.e. in upsampling mode, feeding \OursModuleb{} with a resolution  equivalent to $F^4$. This configuration tests the capability of \OursModule{} to fuse multiscale information while increasing the output resolution.

Based on the computational budget, $F^3$ can also be fed to \OursModuleb{} in the above configurations. However, due to the large feature resolution, decoder computations also come into play, given high-resolution images in the MS-COCO benchmark \cite{mscoco}. Hence, although we have analyzed its effect (See supplement), we do not use this configuration. 

%%%%
%%%%
%%%%
\section{Experiments}
\label{sec:exp}
%
%\subsection{Experimental setup}
%
%
%
\noindent\textbf{Dataset and Evaluation Metric.}
Following \cite{detr, dabdetr, dndetr}, we use MS-COCO $2017$ benchmark \cite{mscoco} for evaluation, having $117$k training images and $5$k validation images. We use MS-COCO’s standard evaluation metric of Average Precision (AP) at different thresholds and different object scales.

\noindent\textbf{Implementation Details.}
We plug the proposed \Ours{} into state-of-the-art DN-DETR \cite{dndetr} for all our experimental evaluations, including ablations. However, to showcase generality, we also adapt \Ours{} into other prominent DETR methods, e.g., Conditional DETR \cite{conddetr}, DAB-DETR \cite{dabdetr} etc.% 

We perform experiments in the $50$-epoch setting, widely used for DETRs \cite{dabdetr, dndetr, imfa}. We also show results for the $12$-epoch or $1\times$ \cite{dndetr} setting to demonstrate accelerated convergence due to the improved DETR design in this paper. The base learning for the backbone is set to $1\times10^{-5}$ while for the transformer, it is set to $1 \times 10^{-4}$. For the $12$-epoch schedule, the learning rate is dropped by $0.1$ at $11^{th}$ epoch, whereas it is dropped at $40^{th}$ epoch for the $50$-epoch schedule. We use $8\times$ NVIDIA A$40$ with a batch size of $16$ ($2$ per-GPU) for training. All the ablations are performed at the $1 \times $ setting.

\begin{table*}[t]
\centering

\caption{Comprehensive evaluation of \Ours{} when applied in prominent DETR models under different settings, i.e., training duration, encoder resolution. Our \Ours{} performs better and faster than DETRs operating at high resolution (DC). `R$50$: ResNet$50$ and `R$18$: ResNet$18$ \cite{resnet}. Refer to Sec.\ref{sec:cred_config} for \halfres{}.}
\label{tab:cred_in_detr}
\vspace{-0.75ex}
\arrayrulecolor{white!60!black}
\setlength{\arrayrulewidth}{0.1ex}

%\scriptsize
%\footnotesize
\scriptsize
%\tiny

\resizebox{0.995\linewidth}{!}{
\setlength{\tabcolsep}{4pt} % do not change

\begin{tabular}{l | c c c c | c c c c c c}

\toprule
 \multicolumn{1}{c}{Method} & \#Epochs & \#Params  & \#FLOPs &\#FPS & AP  & AP$_{50}$ & AP$_{75}$ & AP$_{S}$ & AP$_{M}$ & AP$_{L}$
\\

\midrule
\protect\bdota{}~Conditional DETR-R$50$ \cite{conddetr} & $50$ & $44$M  & $90$G & $26$ & $40.9$ & $61.8$ & $43.3$ & $20.8$ & $44.6$ & $59.2$\\ 

\protect\bdota{}~Conditional DETR-DC$5$-R$50$ \cite{conddetr} & $50$ & $44$M  & $195$G & $15$ & $43.8$ & $64.4$ & $46.7$ & $24.0$ & $47.6$ & $60.7$\\ 

\rowcolor{rwclr}
\protect\bdotb{}~Conditional DETR-R$50$ \cite{conddetr} + \Ours{} & $50$ & $45$M  & $\mathbf{100}$\textbf{G} & $\mathbf{25}$ & $\mathbf{44.4}$ & $\mathbf{64.6}$ & $\mathbf{47.8}$ & $\mathbf{25.2}$ & $\mathbf{46.9}$ & $\mathbf{60.7}$ \\

\midrule

\protect\bdota{}~DAB DETR-R$50$ \cite{dabdetr}   & $50$ & $44$M  & $94$G & $25$ & $42.2$ & $63.1$ & $44.7$ & $21.5$ & $45.7$ & $60.3$ \\ 

\protect\bdota{}~DAB DETR-DC-R$50$ \cite{dabdetr} & $50$ & $45$M  & $202$G & $13$ & $44.5$ & $65.1$ & $47.7$ & $25.3$ & $48.2$ & $62.3$ \\ 

\protect\bdota{}~DAB DETR-R$50$ \cite{dabdetr} + IMFA \cite{imfa} & $50$ & $53$M  & $108$G & $18$ & $45.5$ & $65.0$ & $49.3$ & $27.3$ & $48.3$ & $61.6$ \\

\rowcolor{rwclr}
\protect\bdotb{}~DAB DETR-R$50$ \cite{dabdetr} + \Ours{}  & $50$ & $45$M  & $\mathbf{103}$\textbf{G} & $\mathbf{23}$ & $\mathbf{45.4}$ & $\mathbf{64.9}$ & $\mathbf{49.4}$ & $\mathbf{27.0}$ & $\mathbf{48.5}$ & $\mathbf{62.2}$ \\

\midrule
\protect\bdota{}~DN-DETR-R$50$ \cite{dndetr}   & $50$ & $44$M  & $94$G & $25$ & $44.1$ & $64.4$ & $46.7$ & $22.9$ & $48.0$ & $63.4$\\ 

\protect\bdota{}~DN-DETR-DC5-R$50$ \cite{dabdetr} & $50$ & $44$M  & $202$G & $13$ & $46.3$ & $66.4$ & $49.7$ & $26.7$ & $50.0$ & $64.3$\\ 

\rowcolor{rwclr}
\protect\bdotb{}~DN-DETR-R$50$ \cite{dndetr} + \Ours{} & $50$ & $45$M  & $\mathbf{103}$\textbf{G} & $\mathbf{23}$ & $\mathbf{46.2}$ & $\mathbf{65.8}$ & $\mathbf{49.8}$ & $\mathbf{26.8}$ & $\mathbf{50.0}$ & $\mathbf{63.5}$ \\

\midrule 

\multicolumn{11}{c}{\textbf{$\mathbf{12}$ Epoch Schedule}} \\

\midrule
\protect\bdota{}~Conditional DETR-R$50$ \cite{conddetr}  & $12$ & $44$M  & $90$G & $26$ & $32.4$ & $52.1$ & $33.9$ & $14.2$ & $35.2$ & $48.4$\\ 

\rowcolor{rwclr}
\protect\bdotb{}~Conditional DETR-R$50$ + \Ours{} & $12$ & $45$M  & $100$G & $25$ & $\mathbf{36.6}$ & $\mathbf{56.2}$ & $\mathbf{38.7}$ & $\mathbf{18.8}$ & $\mathbf{39.5}$ & $\mathbf{52.6}$ \\

\midrule
\protect\bdota{}~DAB DETR-R$50$ \cite{dabdetr}  & $12$ & $44$M  & $94$G & $25$ & $35.1$ & $55.5$ & $36.7$ & $16.2$ & $38.1$ & $52.5$ \\ 
\protect\bdota{}~DAB DETR-R$50$-DC5 \cite{dabdetr} & $12$ & $44$M  & $202$G & $13$  & $38.0$ & $60.3$ & $39.8$ & $19.2$ & $40.9$ & $55.4$ \\ 

\protect\bdota{}~DAB DETR-R$50$ \cite{dabdetr} + IMFA \cite{imfa} & $12$ & $53$M  & $108$G & $18$ & $37.3$ & $57.9$ & $39.9$ & $20.8$ & $40.7$ & $52.3$ \\ 

\rowcolor{rwclr}
\protect\bdotb{}~DAB DETR-R$50$ \cite{dabdetr} + \Ours{}  & $12$ & $45$M  & $\mathbf{103}$\textbf{G} & $\mathbf{23}$ &  $\mathbf{38.4}$ & $\mathbf{58.4}$ & $\mathbf{41.0}$ & $\mathbf{20.0}$ & $\mathbf{41.8}$ & $\mathbf{53.9}$ \\

\midrule
\protect\bdota{}~DN-DETR-R$50$ \cite{dndetr}   & $12$ & $44$M  & $94$G & $25$ & $38.6$ & $59.1$ & $41.0$ & $17.3$ & $42.4$ & $57.7$ \\ 

\protect\bdota{}~DN-DETR-DC5-R$50$ \cite{dndetr}  & $12$ & $44$M  & $202$G & $13$ & $41.7$ & $61.4$ & $44.1$ & $21.2$ & $45.0$ & $60.2$\\ 

\rowcolor{rwclr}
\protect\bdotb{}~DN-DETR-R$50$ \cite{dndetr} + \Ours{}  & $12$ & $45$M  & $\mathbf{103}$\textbf{G} & $\mathbf{23}$ &  $\mathbf{41.1}$ & $\mathbf{60.6}$ & $\mathbf{44.0}$ & $\mathbf{22.2}$ & $\mathbf{44.1}$ & $\mathbf{58.9}$ \\

\midrule

\protect\bdota{}~DAB DETR-R$18$ \cite{dabdetr} & $12$ & $31$M  & $49$G & $38$ & $29.8$ & $49.0$ & $30.5$ & $10.9$ & $32.5$ & $46.9$\\ 

\protect\bdota{}~DAB DETR-R$18$ \cite{dabdetr} + IMFA \cite{imfa} & $12$ & $40$M  & $61$G & $23$ & $31.2$ & $51.5$ & $32.3$ & $13.0$ & $33.2$ & $49.1$ \\ 

\rowcolor{rwclr}
\protect\bdotb{}~DAB-DETR-R$18$ \cite{dabdetr} + \Ours{} & $12$ & $32$M  & $\mathbf{60}$\textbf{G} & $\mathbf{31}$ & $\mathbf{33.5}$ & $\mathbf{52.0}$ & $\mathbf{35.2}$ & $\mathbf{16.7}$ & $\mathbf{36.0}$ & $\mathbf{46.2}$ \\

\midrule

\protect\bdota{}~DN-DETR-R$18$ \cite{dndetr}   & $12$ & $31$M  & $49$G & $38$ & $32.5$ & $51.6$ & $33.7$ & $13.5$ & $35.1$ & $49.4$\\ 
\rowcolor{rwclr}
\protect\bdotb{}~DN-DETR-R$18$ \cite{dndetr} + \Ours{} & $12$ & $32$M  & $\mathbf{60}$\textbf{G} & $\mathbf{31}$ & $\mathbf{35.0}$ & $\mathbf{54.0}$ & $\mathbf{36.9}$ & $\mathbf{16.3}$ & $\mathbf{37.0}$ & $\mathbf{51.4}$ \\

\midrule 

\multicolumn{11}{c}{\textbf{\halfres{} Configuration}} \\

\midrule

\protect\bdota{}~DAB DETR-R$50$ \cite{dabdetr}  & $12$ & $44$M  & $94$G & $25$ & $35.1$ & $55.5$ & $36.7$ & $16.2$ & $38.1$ & $52.5$ \\ 

\protect\bdota{}~DAB DETR-R$50$ \cite{dabdetr}~\halfres{} & $12$ & $44$M  & $80$G & $26$ & $28.4$ & $48.9$ & $30.0$ & $9.8$ & $31.5$ & $47.0$\\ 

\protect\bdota{}~DAB DETR-R$50$ \cite{dabdetr} + IMFA \cite{imfa}~\halfres{} & $12$ & $44$M  & $96$G & $18$ & $33.0$ & $54.2$ & $34.5$ & $16.1$ & $35.3$ & $46.5$\\ 

\rowcolor{rwclr}
\protect\bdotb{}~DAB-DETR-R$50$ \cite{dabdetr} + \Ours{}~\halfres{} & $12$ & $45$M  & $\mathbf{94}$\textbf{G} & $\mathbf{24}$ & $\mathbf{37.5}$ & $\mathbf{57.9}$ & $\mathbf{40.1}$ & $\mathbf{18.8}$ & $\mathbf{40.7}$ & $\mathbf{53.0}$ \\

\midrule

\protect\bdota{}~DN-DETR-R$50$ \cite{dndetr}   & $12$ & $44$M  & $94$G & $25$ & $38.6$ & $59.1$ & $41.0$ & $17.3$ & $42.4$ & $57.7$ \\ 

\protect\bdota{}~DN-DETR-R$50$ \cite{dndetr}~\halfres{} & $12$ & $44$M  & $80$G & $26$ & $31.5$ & $52.7$ & $31.5$ & $10.8$ & $33.7$ & $52.0$\\ 
\rowcolor{rwclr}
\protect\bdotb{}~DN-DETR-R$50$ \cite{dndetr} + \Ours{}~\halfres{} & $12$ & $45$M  & $\mathbf{94}$\textbf{G} & $\mathbf{2}4$ & $\mathbf{40.0}$ & $\mathbf{59.4}$ & $\mathbf{42.8}$ & $\mathbf{20.7}$ & $\mathbf{43.1}$ & $\mathbf{56.4}$ \\

\midrule

\protect\bdota{}~DN-DETR-R$50$ \cite{dndetr}   & $50$ & $44$M  & $94$G & $25$ & $44.1$ & $64.4$ & $46.7$ & $22.9$ & $48.0$ & $63.4$\\

\protect\bdota{}~DN-DETR-R$50$ \cite{dndetr}~\halfres{} & $50$ & $44$M  & $80$G & $26$ & $39.9$ & $60.1$ & $41.9$ & $19.2$ & $43.5$ & $59.7$\\ 
\rowcolor{rwclr}
\protect\bdotb{}~DN-DETR-R$50$ \cite{dndetr} + \Ours{}~\halfres{} & $50$ & $45$M  & $\mathbf{94}$\textbf{G} & $\mathbf{24}$ & $\mathbf{45.8}$ & $\mathbf{64.9}$ & $\mathbf{49.1}$ & $\mathbf{25.9}$ & $\mathbf{49.1}$ & $\mathbf{62.8}$ \\

\midrule

\protect\bdota{}~DAB DETR-R$18$ \cite{dabdetr} & $12$ & $31$M  & $49$G & $38$ & $29.8$ & $49.0$ & $30.5$ & $10.9$ & $32.5$ & $46.9$\\

\protect\bdota{}~DAB DETR-R$18$ \cite{dabdetr}~\halfres{} & $12$ & $31$M  & $40$G & $39$ & $24.0$ & $43.8$ & $25.2$ & $4.7$ & $27.9$ & $42.0$\\ 

\protect\bdota{}~DAB DETR-R$18$ \cite{dabdetr} + IMFA \cite{imfa}~\halfres{} & $12$ & $40$M  & $50$G & $25$ & $27.8$ & $46.9$ & $28.8$ & $14.1$ & $29.5$ & $29.5$\\ 

\rowcolor{rwclr}
\protect\bdotb{}~DAB-DETR-R$18$ \cite{dabdetr} + \Ours{}~\halfres{} & $12$ & $32$M  & $\mathbf{51}$\textbf{G} & $\mathbf{34}$ & $\mathbf{32.2}$ & $\mathbf{50.9}$ & $\mathbf{34.1}$ & $\mathbf{15.9}$ & $\mathbf{35.2}$ & $\mathbf{44.9}$ \\

\midrule

\protect\bdota{}~DN-DETR-R$18$ \cite{dndetr}   & $12$ & $31$M  & $49$G & $38$ & $32.5$ & $51.6$ & $33.7$ & $13.5$ & $35.1$ & $49.4$\\ 
\protect\bdota{}~DN-DETR-R$18$ \cite{dndetr}~\halfres{}   & $12$ & $31$M  & $40$G & $39$ & $27.0$ & $46.4$ & $26.9$ & $8.3$ & $28.0$ & $45.6$\\ 
\rowcolor{rwclr}
\protect\bdotb{}~DN-DETR-R$18$ \cite{dndetr} + \Ours{}~\halfres{} & $12$ & $32$M  & $\mathbf{51}$\textbf{G}  & $\mathbf{34}$ & $\mathbf{34.2}$ & $\mathbf{53.0}$ & $\mathbf{36.2}$ & $\mathbf{16.0}$ & $\mathbf{36.1}$ & $\mathbf{50.0}$ \\

\bottomrule

\end{tabular}
}

\vspace{-3ex}
\end{table*}
% \end{wraptable}
%

\subsection{Main Results}

\textbf{\Ours{} in DETR: $\mathbf{50}$-Epoch Setting.} We plug \Ours{} into representative DETR frameworks. Table~\ref{fig:cred_in_detr} shows that \Ours{} boosts the AP in each DETR pipeline. Compared with the baseline without DC, \Ours{} introduces a slight overhead of $9$G FLOPs with only $1-2$FPS drop. Compared with recent IMFA \cite{imfa}, the overhead is $\sim36\%$ less with an improvement of $+4$FPS. Our \Ours{} also performs better regarding runtime than the recent sparse sampling-based method \cite{imfa}.

Further, when plugged in Conditional-DETR \cite{conddetr}, \Ours{} delivers the same accuracy at $50\%$ more FPS than the advanced high-resolution DAB-DETR-DC5-R$50$ \cite{dndetr}. Similarly, with DN-DETR, \Ours{} delivers the same accuracy at $76\%$ more FPS and $50\%$ fewer FLOPs. 

\textbf{\Ours{} in DETR: $\mathbf{12}$-Epoch Setting.} Evaluations in this setting show that \Ours{} speeds up convergence with slight overhead. From Table~\ref{tab:cred_in_detr}, \Ours{} with DAB-DETR \cite{dabdetr} is better than vanilla DAB-DETR by $3.3$AP with only a drop of $2$FPS. Compared with high-resolution DAB-DETR-DC$5$, \Ours{} is accurate by $0.4$AP at $50\%$ fewer FLOPs and $76\%$ more FPS.

\begin{figure}[!t]
\centering

\FPeval{\scal}{0.16}
\begin{tikzpicture}

\colorlet{clr}{white!100!gray}
\node (1)[draw=none, xshift=0ex, yshift=0ex]{
\includegraphics[scale=0.45]{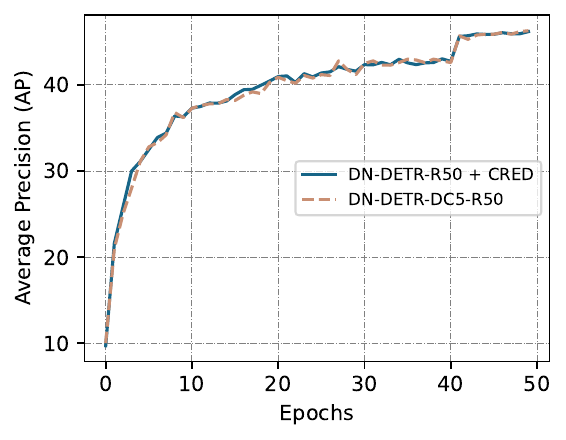}
};
\node (2)[draw=none, xshift=45ex, yshift=0ex]{
\includegraphics[scale=0.45]{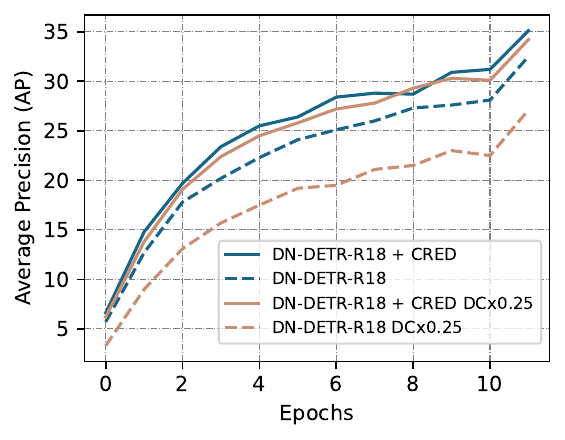}
};
\node (a) [below of=1, xshift=-18ex, yshift=-2ex, scale=0.9]{(a)};
\node (b) [below of=2, xshift=-18ex, yshift=-2ex, scale=0.9]{(b)};

\end{tikzpicture}

\vspace{-2ex}

\caption{Convergence plots over MS-COCO validation set. (a) It can be seen that despite having $50\%$ fewer FLOPs and $76\%$ higher FPS, \Ours{} converges similarly to the baseline. (b) DETR with smaller backbone and their \halfres{} variants in $12$ epoch setting. Notice that even in the smaller backbone, \Ours{}-enabled model with and without DC$\times0.25$ have similar accuracy, but this gap is noticeable in the baselines with and without DC$\times0.25$. This strengthens the utility of \Ours{} that encoder input resolution can be aggressively dropped to save computations while having better accuracy.}

\label{fig:convergence_plot}

\vspace{-4ex}

\end{figure}

With DN-DETR \cite{dndetr}, \Ours{} achieves beyond $41$AP in just $103$G FLOPs, implying that \Ours{} improves the convergence speed (See Figure~\ref{fig:convergence_plot}), i.e. DN-DETR via \Ours{} achieves the performance of its DC counterpart in just $12$ epochs at $50\%$ fewer FLOPs and $76\%$ higher FPS. From the table, \Ours{} can improve the performance of smaller backbones like ResNet$18$ while delivering real-time performance ($>30$FPS). This indicates the utility of \Ours{} where smaller backbones are used due to resource constraints. Hence, by using \Ours{}, detection performance can be boosted

\textbf{\Ours{} in DETR: \halfres{} Setting.}
This setting is crucial to show the utility of \Ours{} in DETRs for real-time performance. From Table~\ref{tab:cred_in_detr}, when encoder resolution is dropped to half (\halfres{}) of any vanilla DETR pipeline, it degrades the performance while also reducing FLOP requirement. However, the degradation in the performance supersedes the reduced FLOPs. 

Whereas \Ours{} in this configuration delivers performance better than the vanilla variant. For example, for vanilla DAB-DETR-R$50$, the AP drops from $35.1$ to $28.4$ with \halfres{}; however, by using \Ours{} in \halfres{} configuration, we achieve $+2.4$AP than the vanilla DAB-DETR-R$50$ in same FLOPs and same FPS. A similar case applies to DN-DETR variants with different backbones.

\begin{table*}[t]
\centering

\caption{Comparison with state-of-the-art object detectors on COCO val $2017$. `MS': Multiscale, `SM': Sparse Multiscale Sampling, and `DC': Dilated Convolution. FPS is reported at $(800\times1280)$ }
\label{tab:cred_vs_sota}
\vspace{-0.75ex}
\arrayrulecolor{white!60!black}
%\setlength{\arrayrulewidth}{0.1ex}

%\scriptsize
%\footnotesize
\scriptsize
%\tiny

\resizebox{1.0\linewidth}{!}{
\setlength{\tabcolsep}{3pt} % do not change

\begin{tabular}{l | c c c | r r r c | c c c c c c}

\toprule
  \multicolumn{1}{c}{Method} & MS & SM & DC & \#Epochs & \#Params  & \#FLOPs &\#FPS & AP  & AP$_{50}$ & AP$_{75}$ & AP$_{S}$ & AP$_{M}$ & AP$_{L}$
\\

\midrule
\protect\bdota{}~YOLOS-DeiT-S \cite{yolodeit}   & & &  & $150$ & $28$M & $172$G & $-$ &  $37.6$ & $57.6$ & $39.2$ & $15.9$ & $40.2$ & $57.3$ \\
\protect\bdota{}~Faster-RCNN-FPN-R$50$ \cite{fasterrcnn, fpn} & \cmark & & & $108$ & $42$M & $180$G & $-$ &  $42.0$ & $62.1$ & $45.5$ & $26.6$ & $45.5$ & $53.4$ \\
\protect\bdota{}~TSP-FCOS-FPN-R$50$ \cite{tspfcos}  & \cmark & & & $36$  & $52$M & $189$G & $-$ & $43.1$ & $62.3$ & $47.0$ & $26.6$ & $46.8$ & $55.9$ \\
\protect\bdota{}~TSP-RCNN-FPN-R$50$  \cite{tspfcos}   & \cmark & & & $36$ & $64$M & $188$G & $-$ &  $43.8$ & $63.3$ & $48.3$ & $28.6$ & $46.9$ & $55.7$ \\
\protect\bdota{}~Faster RCNN-FPN-R$101$ \cite{fasterrcnn, fpn} & \cmark & &  & $108$ & $60$M & $246$G & $-$ & $44.0$ & $63.9$ & $47.8$ & $27.2$ & $48.1$ & $56.0$ \\
\protect\bdota{}~Sparse-RCNN-FPN-R$50$ \cite{sparsercnn}   & \cmark & & & $36$ & $106$M & $166$G & $-$ &  $45.0$ & $64.1$ & $48.9$ & $28.0$ & $47.6$ & $59.5$ \\

\midrule
\protect\bdota{}~DETR-DC5-R50 \cite{detr}  & & & \cmark &  $500$ & $41$M & $187$G & $16$ &  $43.3$ & $63.1$ & $45.9$ & $22.5$ & $47.3$ & $61.1$ \\
\protect\bdota{}~SAM-DETR-DC5-R$50$ \cite{samdetr}   & & & \cmark &  $50$ & $58$M & $210$G & $-$ &  $43.3$ & $64.4$ & $46.2$ & $25.1$ & $46.9$ & $61.0$ \\
\protect\bdota{}~SMCA-DETR-R$50$ \cite{smcadetr}   & \cmark & &  &  $50$ & $40$M & $152$G & $-$ &  $43.7$ & $63.6$ & $47.2$ & $24.2$ & $47.0$ & $60.4$ \\ 
\protect\bdota{}~Deformable-DETR-R$50$ \cite{deformabledetr}   & \cmark & &  &  $50$ & $40$M & $173$G & $13$ &  $43.8$ & $62.6$ & $47.7$ & $26.4$ & $47.1$ & $58.0$ \\
\protect\bdota{}~Conditional-DETR-DC5-R$50$ \cite{conddetr}   & & & \cmark &  $50$ & $44$M & $195$G & $15$ &  $43.8$ & $64.4$ & $46.7$ & $24.0$ & $47.6$ & $60.7$ \\
\protect\bdota{}~Anchor-DETR-DC5-R$50$ \cite{anchordetr}   & & & \cmark &  $50$ & $37$M & $172$G & $-$ &  $44.2$ & $64.7$ & $47.5$ & $24.7$ & $48.2$ & $60.6$ \\
\protect\bdota{}~Efficient-DETR-R$50$ \cite{efficientdetr}   & \cmark & &  &  $36$ & $32$M & $159$G & $-$ &  $44.2$ & $62.2$ & $48.0$ & $28.4$ & $47.5$ & $56.6$ \\
\protect\bdota{}~DAB-DETR-DC5-R$50$ \cite{dabdetr}   & & & \cmark &  $50$ & $44$M & $202$G & $13$ &  $44.5$ & $65.1$ & $47.7$ & $25.3$ & $48.2$ & $62.3$ \\
\protect\bdota{}~SAM-DETR-DC5-R$50$ \cite{samdetr} w/ SMCA   & & & \cmark &  $50$ & $58$M & $210$G & $-$ &  $45.0$ & $65.4$ & $47.9$ & $26.2$ 
& $49.0$ & $63.3$ \\

\protect\bdota{}~Conditional DETR-DC5-R$101$ \cite{conddetr} & & & \cmark   & $50$ & $63$M  & $262$G & $10$ & $45.0$ & $65.5$ & $48.4$ & $26.1$ & $48.9$ & $62.8$ \\ 

\protect\bdota{}~DN-DETR-R$101$ \cite{dndetr} & & & \cmark &  $50$ & $63$M & $174$G & $-$ & $45.2$ & $65.5$ & $48.3$ & $24.1$ & $49.1$ & $65.1$ \\

\protect\bdota{}~Deformable-DAB-DETR-R$50$ \cite{deformabledetr}   & \cmark & \cmark &  &  $50$ & $41$M & $173$G & $12$ &  $45.4$ & $64.7$ & $49.0$ & $26.8$ & $48.3$ & $61.7$ \\

\protect\bdota{}~IMFA-DAB-DETR-R$50$ \cite{imfa}   & & \cmark &  &  $50$ & $53$M & $108$G & $18$ &  $45.5$ & $65.0$ & $49.3$ & $27.3$ & $48.3$ & $61.6$ \\

\protect\bdota{}~DAB DETR-DC5-R$101$ \cite{dabdetr} & & & \cmark & $50$ & $63$M  & $282$G  & $10$ & $45.8$ & $65.9$ & $49.3$ & $27.0$ & $49.8$ & $63.3$ \\

\protect\bdota{}~SAP-DETR-DC5-R$50$ \cite{sapdetr} & & & & $50$ & $47$M  & $197$G & $12$ & $46.0$ & $65.5$ & $48.9$ & $26.4$ & $50.2$ & $62.6$\\ 

\protect\bdota{}~DN-DETR-DC5-R$50$ \cite{dndetr} & & & \cmark & $50$ & $44$M  & $202$G & $13$ & $46.3$ & $66.4$ & $49.7$ & $26.7$ & $50.0$ & $64.3$\\ 

\protect\bdota{}~Siamese-DETR-R$50$ \cite{siamesedetr} & \cmark & \cmark &  & $50$ & $41$M  & $173$G & $-$ & $46.3$ & $64.6$ & $50.5$ & $28.1$ & $50.1$ & $61.5$\\ 

\protect\bdota{}~Lite-DETR-R$50$ \cite{litedetr} & \cmark & \cmark & & $50$ & $41$M  & $123$G & $15$ & $46.7$ & $66.1$ & $50.6$ & $29.1$ & $49.7$ & $62.2$\\ 

\protect\bdota{}~SAP-DETR-DC5-R$101$ \cite{sapdetr} & & & & $50$ & $67$M  & $266$G & $11$ & $46.9$ & $66.7$ & $50.5$ & $27.9$ & $51.3$ & $64.3$\\

\midrule 

\rowcolor{rwclr}
\protect\bdotb{}~DAB DETR-R$50$ \cite{dabdetr} + \Ours{}  & & & & $50$ & $45$M  & $\mathbf{103}$\textbf{G} & $\mathbf{23}$ & $\mathbf{45.4}$ & $\mathbf{64.9}$ & $\mathbf{49.4}$ & $\mathbf{27.0}$ & $\mathbf{48.5}$ & $\mathbf{62.2}$ \\

\rowcolor{rwclr}
\protect\bdotb{}~DN-DETR-R$50$ \cite{dndetr} + \Ours{}~\halfres{} & & &  & $50$ & $45$M  & $\mathbf{94}$\textbf{G} & $\mathbf{24}$ & $\mathbf{45.8}$ & $\mathbf{64.9}$ & $\mathbf{49.1}$ & $\mathbf{25.9}$ & $\mathbf{49.1}$ & $\mathbf{62.8}$ \\

\rowcolor{rwclr}
\protect\bdotb{}~DN-DETR-R$50$ \cite{dndetr} + \Ours{} & & &  & $50$ & $45$M  & $\mathbf{103}$\textbf{G} & $\mathbf{23}$ & $\mathbf{46.2}$ & $\mathbf{65.8}$ & $\mathbf{49.8}$ & $\mathbf{26.8}$ & $\mathbf{50.0}$ & $\mathbf{63.5}$ \\

\rowcolor{rwclr}
\protect\bdotb{}~DN-DETR-R$50$ \cite{dndetr} + \Ours{}-OO & & &  & $50$ & $45$M  & $\mathbf{105}$\textbf{G} & $\mathbf{23}$ & $\mathbf{46.8}$ & $\mathbf{66.8}$ & $\mathbf{50.5}$ & $\mathbf{27.4}$ & $\mathbf{50.7}$ & $\mathbf{64.0}$ \\

\midrule

\multicolumn{14}{c}{\textbf{$\mathbf{12}$ Epoch Schedule}} \\

\midrule

\protect\bdota{}~DETR-R$50$ \cite{detr} &  &  &  & $12$ & $41$M  & $86$G & $27$ & $15.5$ & $29.4$ & $14.5$ & $4.3$ & $15.1$ & $26.7$\\

\protect\bdota{}~Deformable DETR-R$50$ \cite{deformabledetr} &  & \cmark &  & $12$ & $40$M  & $173$G & $12$ & $37.2$ & $55.5$ & $40.5$ & $21.1$ & $40.7$ & $50.5$\\

\protect\bdota{}~DAB DETR-R$50$ \cite{dabdetr} + IMFA \cite{imfa} & & \cmark &  & $12$ & $53$M  & $108$G & $18$ & $37.3$ & $57.9$ & $39.9$ & $20.8$ & $40.7$ & $52.3$ \\ 

\protect\bdota{}~DAB DETR-DC-R$101$ \cite{detr} & &  & \cmark  & $12$ & $63$M  & $282$G & $10$ & $40.3$ & $62.6$ & $42.7$ & $22.2$ & $44.0$ & $57.3$\\ 

\midrule

\rowcolor{rwclr}
\protect\bdotb{}~DN-DETR-R$18$ \cite{dndetr} + \Ours{}~\halfres{} & & &  & $12$ & $32$M  & $\mathbf{51}$\textbf{G}  & $\mathbf{34}$ & $\mathbf{34.2}$ & $\mathbf{53.0}$ & $\mathbf{36.2}$ & $\mathbf{16.0}$ & $\mathbf{36.1}$ & $\mathbf{50.0}$ \\

\rowcolor{rwclr}
\protect\bdotb{}~DN-DETR-R$18$ \cite{dndetr} + \Ours{} & & &  & $12$ & $32$M  & $\mathbf{60}$\textbf{G} & $\mathbf{31}$ & $\mathbf{35.0}$ & $\mathbf{54.0}$ & $\mathbf{36.9}$ & $\mathbf{16.3}$ & $\mathbf{37.0}$ & $\mathbf{51.4}$ \\

\rowcolor{rwclr}
\protect\bdotb{}~DAB-DETR-R$50$ \cite{dabdetr} + \Ours{}~\halfres{} & & & & $12$ & $45$M  & $\mathbf{94}$\textbf{G} & $\mathbf{24}$ & $\mathbf{37.5}$ & $\mathbf{57.9}$ & $\mathbf{40.1}$ & $\mathbf{18.8}$ & $\mathbf{40.7}$ & $\mathbf{53.0}$ \\

\rowcolor{rwclr}
\protect\bdotb{}~DAB DETR-R$50$ \cite{dabdetr} + \Ours{} & & &   & $12$ & $45$M  & $\mathbf{103}$\textbf{G} & $\mathbf{23}$ &  $\mathbf{38.4}$ & $\mathbf{58.4}$ & $\mathbf{41.0}$ & $\mathbf{20.0}$ & $\mathbf{41.8}$ & $\mathbf{53.9}$ \\

\rowcolor{rwclr}
\protect\bdotb{}~DN-DETR-R$50$ \cite{dndetr} + \Ours{}~\halfres{} & & &  & $12$ & $45$M  & $\mathbf{94}$\textbf{G} & $\mathbf{24}$ & $\mathbf{40.0}$ & $\mathbf{59.4}$ & $\mathbf{42.8}$ & $\mathbf{20.7}$ & $\mathbf{43.1}$ & $\mathbf{56.4}$ \\

\rowcolor{rwclr}
\protect\bdotb{}~DN-DETR-R$50$ \cite{dndetr} + \Ours{} & & & & $12$ & $45$M  & $\mathbf{103}$\textbf{G} & $\mathbf{23}$ &  $\mathbf{41.1}$ & $\mathbf{60.6}$ & $\mathbf{44.0}$ & $\mathbf{22.2}$ & $\mathbf{44.1}$ & $\mathbf{58.9}$ \\

\bottomrule

\end{tabular}
}
\vspace{-5ex}
\end{table*}
% \end{wraptable}
%

\noindent\textbf{\Ours{} vs State-of-the-art.}
We also compare our \Ours{} with state-of-the-art object detection pipelines with multiscale, high-resolution, and sparse sampling approaches. Table~\ref{tab:cred_vs_sota} shows the results. From the table, it can be seen that \Ours{}-DETR models are better by a large margin ($>50\%$) in FLOPs and FPS while delivering accuracy comparable with state-of-the-art methods. Even ResNet-$18$ based models with \Ours{} show competitive performance with Deformable-DETR \cite{deformabledetr}, DAB-DETR \cite{dabdetr}, IMFA \cite{imfa} with a stronger backbone ResNet-$101$.

\Ours{} w/ ResNet-$50$ performs better than heavy models, even in \halfres{} configuration and $12$-epoch setting. For example, DN-DETR-R$50$ + \Ours{} \halfres{} is better than multiscale Deformable-DETR-R$50$ \cite{deformabledetr} by $2.8$AP, $45\%$ fewer FLOPs and $50\%$ higher FPS. Similarly, DN-DETR-R$50$ + \Ours{} is better than DAB-DETR-DC$5$-R$101$ by $0.8$AP, $63\%$ fewer FLOPs, and $130\%$ higher FPS. Then DN-DETR-R$50$ + \Ours{}-OO has $60\%$ fewer FLOPs than \cite{sapdetr} with the same accuracy. Furthermore, the accuracy can be improved using the latest DETR-training techniques of \cite{msdetr, saliencedetr}, which we leave for future work.

\begin{figure}[!t]

\centering

\begin{tikzpicture}

\colorlet{clr}{white!100!gray}

\node (1)[draw=none, xshift=0ex, yshift=0ex]{
\includegraphics[scale=0.55]{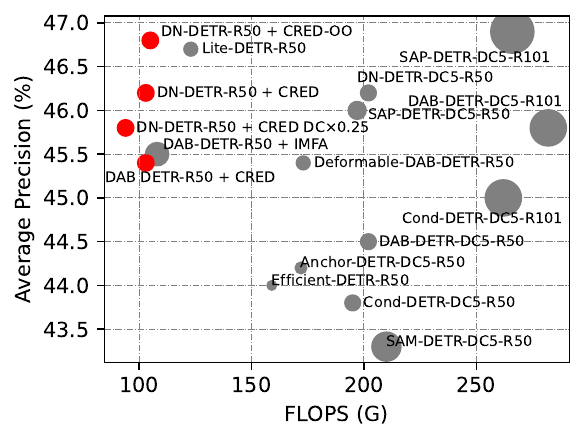}
};

\node (2)[draw=none, xshift=44ex, yshift=0ex]{
\includegraphics[scale=0.55]{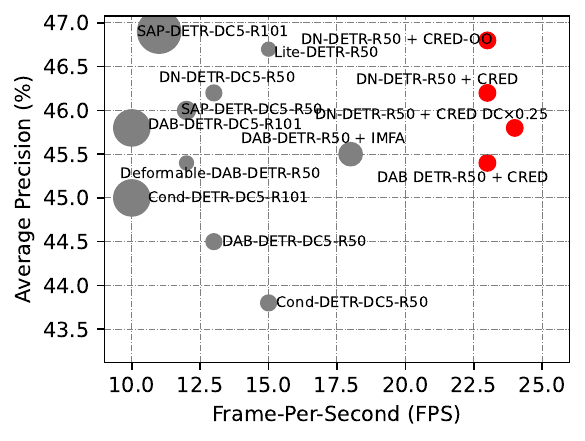}
};

\node (3)[draw=none, xshift=0.5ex, yshift=-26ex]{
\includegraphics[scale=0.55]{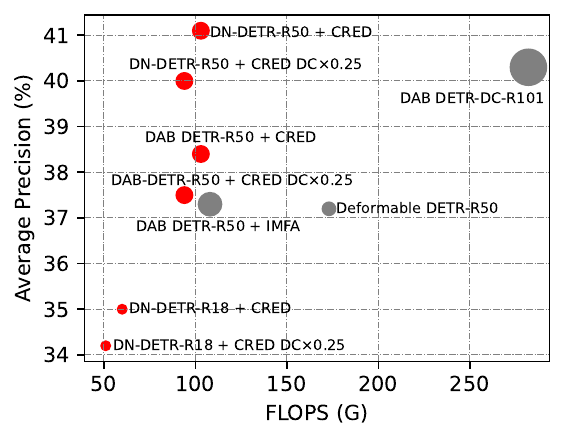}
};

\node (4)[draw=none, xshift=45ex, yshift=-26ex]{
\includegraphics[scale=0.55]{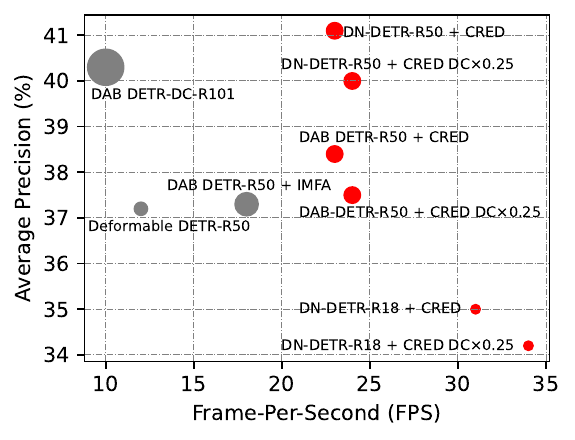}
};

\node (a) [below of=1, xshift=-18ex, yshift=-3.5ex, scale=0.9]{(a)};
\node (b) [below of=2, xshift=-18ex, yshift=-3.5ex, scale=0.9]{(b)};
\node (c) [below of=3, xshift=-18ex, yshift=-3.5ex, scale=0.9]{(c)};
\node (d) [below of=4, xshift=-18ex, yshift=-3.5ex, scale=0.9]{(d)};

\end{tikzpicture}

\vspace{-2.0ex}
%
% %
\def\circ{\protect\raisebox{0.2ex}{\protect\tikz{\protect\node[draw=red, fill=red,scale=0.5, circle]{};}}}
\def\circg{\protect\raisebox{0.2ex}{\protect\tikz{\protect\node[draw=gray, fill=gray,scale=0.5, circle]{};}}}
\caption{The proposed \Ours{} \textit{vs} representative DETRs \cite{dabdetr, dndetr, deformabledetr, sapdetr, litedetr} etc. (a,b) $50$-Epochs, and (c,d) $12$-epochs. `\protect\circ{} and `\protect\circg{} refers to \Ours{} and existing models respectively. The size of the circle is proportional to the parameter count.}

\label{fig:cred_vs_sota}
\vspace{-2ex}

\end{figure}
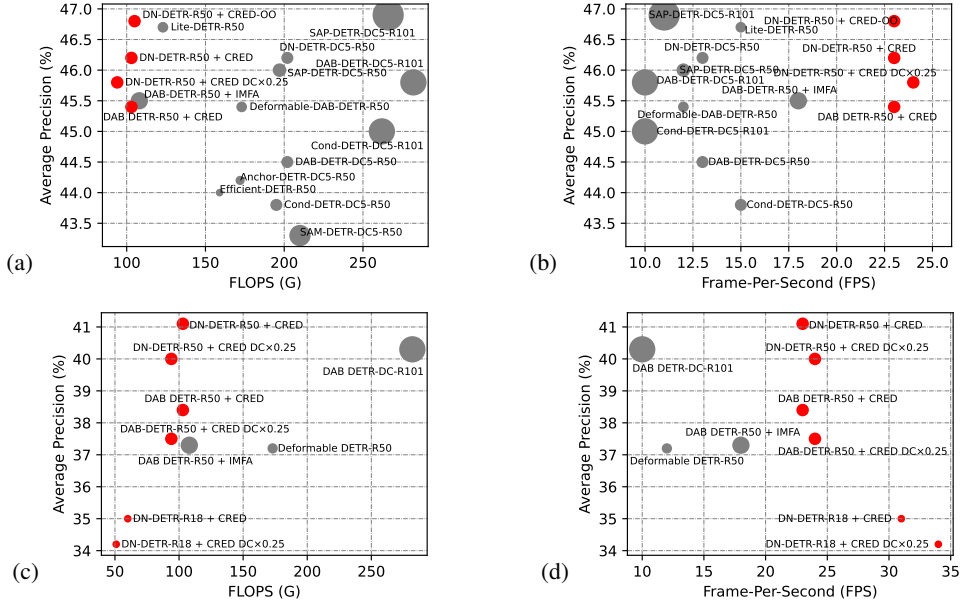

Figure~\ref{fig:cred_vs_sota} further strengthens our results, that \Ours{}, while delivering comparable performance to the state-of-the-art, have far fewer FLOPs and higher FPS. Also, the results indicate that by utilizing the \halfres{} configuration in \Ours{}, DETRs of real-time speed and high accuracy can be constructed, indicating the huge potential of \OursFull{} in state-of-the-art DETRs.

\subsection{Ablations}
\label{sec:ablations}

We conduct a comprehensive ablation study on \Ours{} design by using the state-of-the-art DN-DETR \cite{dndetr} framework and provide insight on the design motivations.

\begin{table}[t]

\centering

%\vspace{1ex}

\captionof{table}{Effect of \OursModule{} and \OursModuleb{} in \Ours{}.}
\label{tab:ablation_osma_cram}
\vspace{-0.75ex}
\arrayrulecolor{white!60!black}
%\setlength{\arrayrulewidth}{0.1ex}

%\scriptsize
%\footnotesize
%\scriptsize
%\tiny

\resizebox{0.55\linewidth}{!}{
\setlength{\tabcolsep}{3pt}

\begin{tabular}{c c | c c | c c c c c c}

\toprule
 \OursModule & \OursModuleb{} & \#Params  & \#FLOPs & AP  & AP$_{50}$ & AP$_{75}$ & AP$_{S}$ & AP$_{M}$ & AP$_{L}$ \\
\midrule

 &  & $44$M  & $94$G & $38.6$ & $59.1$ & $41.0$ & $17.3$ & $42.4$ & $57.7$ \\
\cmark &  & $45$M  & $97$G & $39.8$ & $59.7$ & $42.5$ & $19.1$ & $43.2$ & $58.1$ \\
 & \cmark  & $45$M  & $100$G & $40.4$ & $60.1$ & $43.1$ & $21.2$ & $43.6$ & $58.3$ \\
\rowcolor{rwclr}
\cmark & \cmark  & $45$M  & $103$G & $41.1$ & $60.6$ & $44.0$ & $22.2$ & $44.1$ & $58.9$  \\

\bottomrule

\end{tabular}
}
\vspace{-1ex}
\end{table}

\begin{table}[!h]
\centering

\vspace{-0.75ex}

\captionof{table}{Computation complexity analysis of \Ours{}.}
\label{tab:cred_complexity_analysis}
\arrayrulecolor{white!60!black}
%\setlength{\arrayrulewidth}{0.1ex}

%\scriptsize
%\footnotesize
% \scriptsize
%\tiny

\resizebox{0.8\linewidth}{!}{
\setlength{\tabcolsep}{3pt}

\begin{tabular}{l | r r r r r | r | c c}

\toprule
 \multicolumn{1}{c}{Method} & Backbone & Encoder  & Decoder & \OursModuleb{} &  \OursModule{} & Total & FPS & AP$_{50}$ \\

\midrule

\protect\bdota{}~DN-DETR-R$50$ \cite{dndetr} & $74$G & $12$G & $8$G & $-$  & $-$ & $94$G & $25$ & $44.1$ \\
\protect\bdota{}~DN-DETR-DC$5$-R$50$ \cite{dndetr} & $112$G & $80$G & $10$G & $-$  & $-$   & $202$G & $13$ & $46.3$ \\
\protect\bdotb{}~DN-DETR-R$50$ \cite{dndetr} + \Ours{} & $74$G & $12$G & $10$G & $3$G & $2$G & $103$G & $23$ & $46.2$ \\

\bottomrule

\end{tabular}
}

\vspace{-3ex}

\end{table}

\begin{table}[t]

\centering 
\captionof{table}{Ablation of \OursModule{} design.}
\label{tab:ablation_osma}
\vspace{-0.75ex}
\arrayrulecolor{white!60!black}
%\setlength{\arrayrulewidth}{0.1ex}

%\scriptsize
%\footnotesize
%\scriptsize
%\tiny

\resizebox{0.68\linewidth}{!}{
\setlength{\tabcolsep}{5pt}

\begin{tabular}{c | c c c | c c c | c c | c c c c c c}

\toprule
 ablation & $F^5$ &  $F^4$ & $F^3$ & $g_0$ & $P$  & $d$ & \#Params  & \#FLOPs & AP  & AP$_{50}$ & AP$_{75}$ & AP$_{S}$ & AP$_{M}$ & AP$_{L}$ \\
\midrule

 & \cmark & \cmark &  & $1$ & $1$ & $21$ & $45$M  & $101$G & $40.2$ & $59.5$ & $43.2$ & $21.3$ & $43.2$ & $57.9$ \\
 \rowcolor{rwclr}
 \multirow{-2}{*}{\shortstack{changing \\ \#scales}} & 
\cmark & \cmark & \cmark & $1$ & $1$ & $21$ & $45$M  & $103$G & $41.1$ & $60.6$ & $44.0$ & $22.2$ & $44.1$ & $58.9$  \\

\midrule

 & \cmark & \cmark & \cmark & $2$ & $2$ & $21$ & $45$M  & $101$G & $39.7$ & $59.5$ & $42.3$ & $21.0$ & $42.7$ & $56.8$ \\

\multirow{-2}{*}{\shortstack{Vary `$\{g_0,P\}$' \\ Same res.}} & \cmark & \cmark & \cmark & $1$ & $1$ & $21$ & $45$M  & $103$G & $41.1$ & $60.6$ & $44.0$ & $22.2$ & $44.1$ & $58.9$ \\

\midrule

 & \cmark & \cmark & \cmark & $2$ & $1$ & $21$ & $45$M  & $94$G & $40.0$ & $59.4$ & $42.8$ & $20.7$ & $43.1$ & $56.4$  \\

\multirow{-2}{*}{\shortstack{Vary `$\{g_0,P\}$' \\ \halfres{}.}} & \cmark & \cmark & \cmark & $4$ & $4$ & $21$ & $45$M  & $92$G & $39.4$ & $59.0$ & $42.1$ & $19.8$ & $42.7$ & $56.0$ \\

\midrule

Vary `$d$' & \cmark & \cmark & \cmark & $1$ & $1$ & $40$ & $45$M  & $114$G & $41.5$ & $61.1$ & $44.2$ & $22.4$ & $44.8$ & $59.6$ \\

\bottomrule

\end{tabular}
}

\vspace{-2ex}

\end{table}

\noindent\textbf{Effect of \OursModuleb{} and \OursModule{}.}
We analyze the effect of \OursModule{} (Sec.~\ref{sec:osma}) and \OursModuleb{} (Sec.~\ref{sec:cram}). Table~\ref{tab:ablation_osma_cram} shows the analysis. It can be seen that by using any of \OursModule{} or \OursModuleb{} into the baseline, the accuracy improves. By using \OursModule{} alone, AP increases by $1.2$, indicating that \OursModule{} produces better input features or tokens for the encoder. While by using only \OursModuleb{}, AP improves by $1.8$AP. Using both the modules, we get an overall improvement of $2.5$AP at a reduction of only $2$FPS and negligible parameter overhead. Interestingly, AP for small objects increases by $\sim5$AP. If we compare this with the DC variant of the baseline, we perform on par in roughly $99$G fewer FLOPs, which is inspirational. This accuracy gap is reduced in the $50$ epoch setting.

Hence, we conjecture that feeding multiscale information to the encoder via \OursModule{} while transferring the low-resolution encoded information to high-resolution feature via \OursModuleb{} proves to be highly effective in DETRs from both accuracy and runtime perspective.

\noindent\textbf{Computational Complexity.}
Table~\ref{tab:cred_complexity_analysis} shows the effect of using \OursModuleb{} and \OursModule{} in DN-DETR. It can be seen that the overhead of these modules is $\sim 5$G FLOPs. However, overall computations slightly increase due to the increased resolution of the decoder when fed by \OursModuleb{}.

\noindent\textbf{Ablation of \OursModule{} Hyperparameters.}
\OursModule{} mainly has intermediate projection dimension $d$ and grid size $g$ as hyperparameters. Another ablation exists within the \OursModule{} design, i.e., the number of input multiscale features. We study them individually in Table~\ref{tab:ablation_osma}.

The first two rows show the effect of input multiscale features. As we include more high-resolution features, overall AP increases along with the AP of small objects. This indicates \OursModule{}'s one-step attention mechanism effectively produces feature maps infused with multiscale information.

We also change grid size $g$. Increasing the grid size $g_0=2$ reduces the FLOPs by $2$G because $T$ (Figure~\ref{fig:osma_step1}) increases and $N_g$ decreases. However, we observed a reduction in the AP. We hypothesize that this happens because the stage-$5$ ($F^5$) feature is the smallest resolution. When more than two features using $g_0=2$ are fused with high-res features, the individual feature at low-resolution loses its chance to interact with the high-resolution features individually because these features already have relatively large receptive fields and carry more information. 

Although we are interested in keeping the values of $d$ equal to $T$, we analyze its effect. We observe that it increases the FLOPs while slightly improving the AP. Hence, based on the computational budget requirements, one can change $d$ to achieve the desired performance and runtime.

\begin{table}[h]
\centering

\vspace{-0.75ex}

\caption{Effect of LayerNorm \cite{layernorm} and activations in \Ours{}.}
\label{tab:ablation_cred}
\arrayrulecolor{white!60!black}
%\setlength{\arrayrulewidth}{0.1ex}

%\scriptsize
%\footnotesize
% \scriptsize
%\tiny

\resizebox{0.6\linewidth}{!}{
\setlength{\tabcolsep}{4pt}

\begin{tabular}{c c | c c | c c c c c c}

\toprule
 LayerNorm & Activation & \#Params  & \#FLOPs & AP  & AP$_{50}$ & AP$_{75}$ & AP$_{S}$ & AP$_{M}$ & AP$_{L}$ \\
\midrule

\cmark & ReLU  & $45$M  & $103$G & $40.6$ & $60.4$ & $43.6$ & $21.6$ & $43.6$ & $57.9$ \\
\rowcolor{rwclr}
\cmark & SiLU & $45$M  & $103$G & $41.1$ & $60.6$ & $44.0$ & $22.2$ & $44.1$ & $58.9$  \\
\xmark & SiLU  &  $45$M  & $103$G & $39.3$ & $59.1$ & $41.9$ & $20.8$ & $42.5$ & $56.6$ \\

\bottomrule

\end{tabular}
}

%\vspace{-1ex}
%
\end{table}
% \end{wraptable}
%

\begin{table}[!h]
\centering

\vspace{-0.75ex}

\caption{Ablation of \OursModuleb{} design.}
\label{tab:ablation_cram}
%
%\vspace{-0.75ex}
%
\arrayrulecolor{white!60!black}
%\setlength{\arrayrulewidth}{0.1ex}

%\scriptsize
%\footnotesize
% \scriptsize
%\tiny

\resizebox{0.5\linewidth}{!}{
\setlength{\tabcolsep}{4pt}

\begin{tabular}{c | c c | c c c c c c}

\toprule
 Input & \#Params  & \#FLOPs & AP  & AP$_{50}$ & AP$_{75}$ & AP$_{S}$ & AP$_{M}$ & AP$_{L}$ \\
\midrule

$F^4$ & $45$M  & $103$G & $41.1$ & $60.6$ & $44.0$ & $22.2$ & $44.1$ & $58.9$ \\
$F^3$ & $45$M  & $147$G & $42.1$ & $61.4$ & $44.9$ & $23.3$ & $44.8$ & $59.4$  \\

\bottomrule

\end{tabular}
}
%\vspace{-1ex}
%
\end{table}
% \end{wraptable}
%

\noindent\textbf{\Ours{} Design.}
Within the \Ours{} design, we study the effect of different activations and the specified use of layer normalization \cite{layernorm}. Table~\ref{tab:ablation_cred} shows that using ReLU or removing LayerNorm from \Ours{} decreases accuracy. This justifies the configuration described in the paper.

\noindent\textbf{Ablation of \OursModuleb{}.}
\OursModuleb{} is studied by changing its input resolution and source. Table~\ref{tab:ablation_cram} shows that despite feeding the encoder with low resolution, \OursModuleb{} can effectively transfer the encoder knowledge to the high-resolution feature. By default, we feed resolution equal to $F^4$ to \OursModuleb{}. When we feed \OursModuleb{} with $F^3$, the computations in the decoder increase mainly in the cross-attention. Although it improves AP, the rise in FLOPs is notable. Hence, we restrict ourselves to feeding \OursModuleb{} with resolution up to $F^4$.

%%%%
%%%%
%%%%
\section{Conclusion}
\label{sec:conc}
In this work, we present a novel \OursFull{} (\Ours{}) mechanism to improve the accuracy and runtime of DETR methods. \Ours{} is based on its two novel modules  \OursModuleFullb{} (\OursModuleb{}) and \OursModuleFull{} (\OursModule{}). \OursModuleb{} transfers the knowledge of low-resolution encoder output to a high-resolution feature. While \OursModule{} is designed to fuse multiscale features in a single step and produce a feature map of a desired resolution. With the application of \Ours{} into state-of-the-art DETR methods, FLOPs get reduced by $50\%$, and FPS increases by $76\%$ than the high-resolution DETR at equivalent detection performance.

\noindent\textbf{Future Scope \& Limitations:} \Ours{} with its promising results shows huge potential in real-time and affordable DETRs with high accuracy and high-resolution image processing. There is greater scope for improvements, e.g., fusing \OursModuleb{} and \OursModule{} for even higher performance or adapting \Ours{} to sparse sampling-based DETRs because the current design can not fuse high-resolution features with sparsely sampled encoder embeddings. In addition, \Ours{} has huge scope in Transformer-based semantic or instance segmentation by leveraging its attention transfer to improve runtime for processing high-resolution images because semantic segmentation produces high-resolution outputs.

%%%%
%%%%
%%%%

\bibliographystyle{iclr2025_conference}
\bibliography{bibfile}

\appendix

\section{Detection Visualizations on MS-COCO Validation Set}

\begin{figure}[!h]
% \begin{wrapfigure}{r}{0.40\textwidth}

\centering

\FPeval{\imw}{30}
\FPeval{\imh}{20}

\begin{tikzpicture}

\node (im1) [xshift=0ex, yshift=0*(\imh ex+0.75ex)]{\includegraphics[width=\imw ex, height=\imh ex]{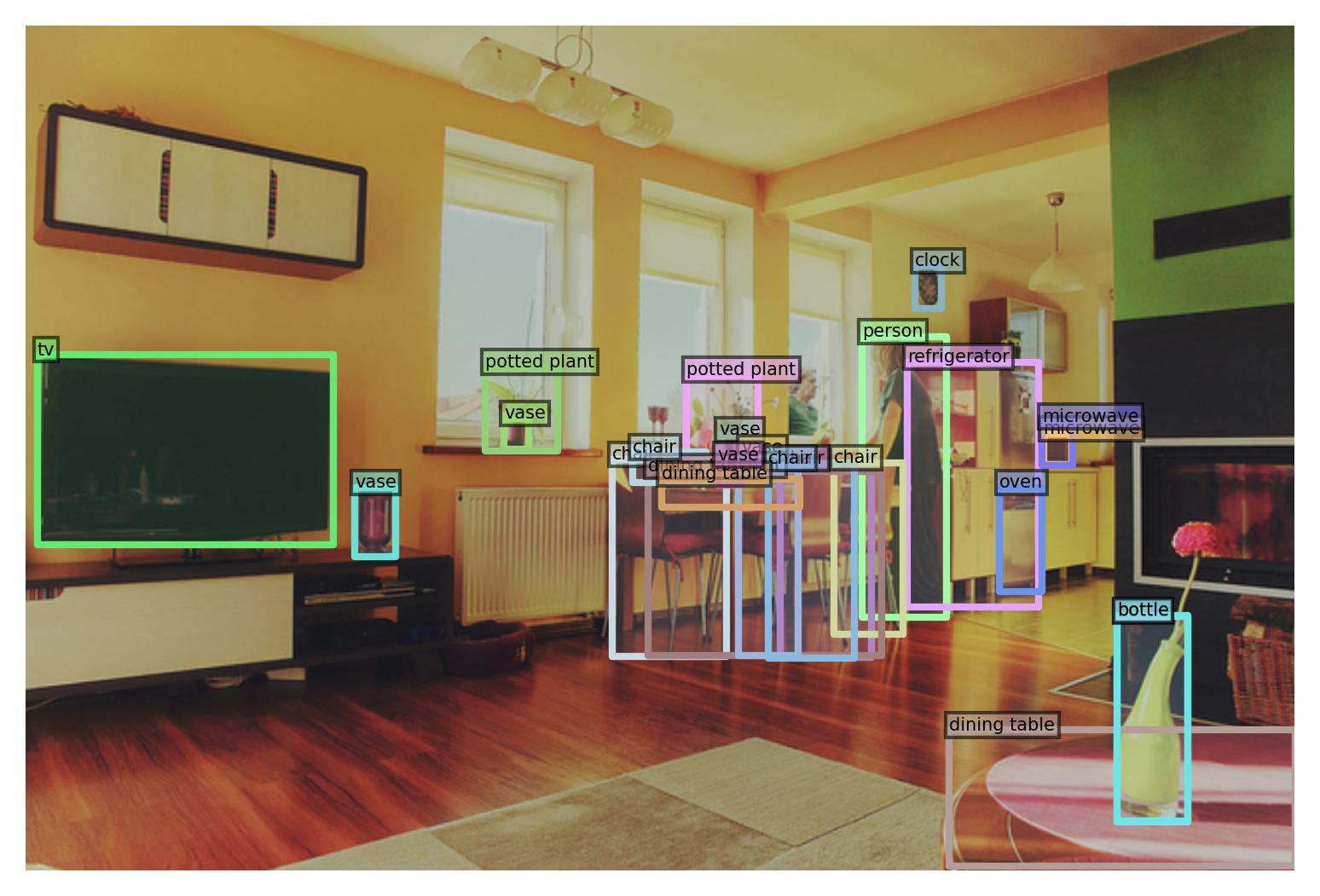}};

\node (im2) [xshift=\imw ex, yshift=0*(\imh ex+0.75ex)]{\includegraphics[width=\imw ex, height=\imh ex]{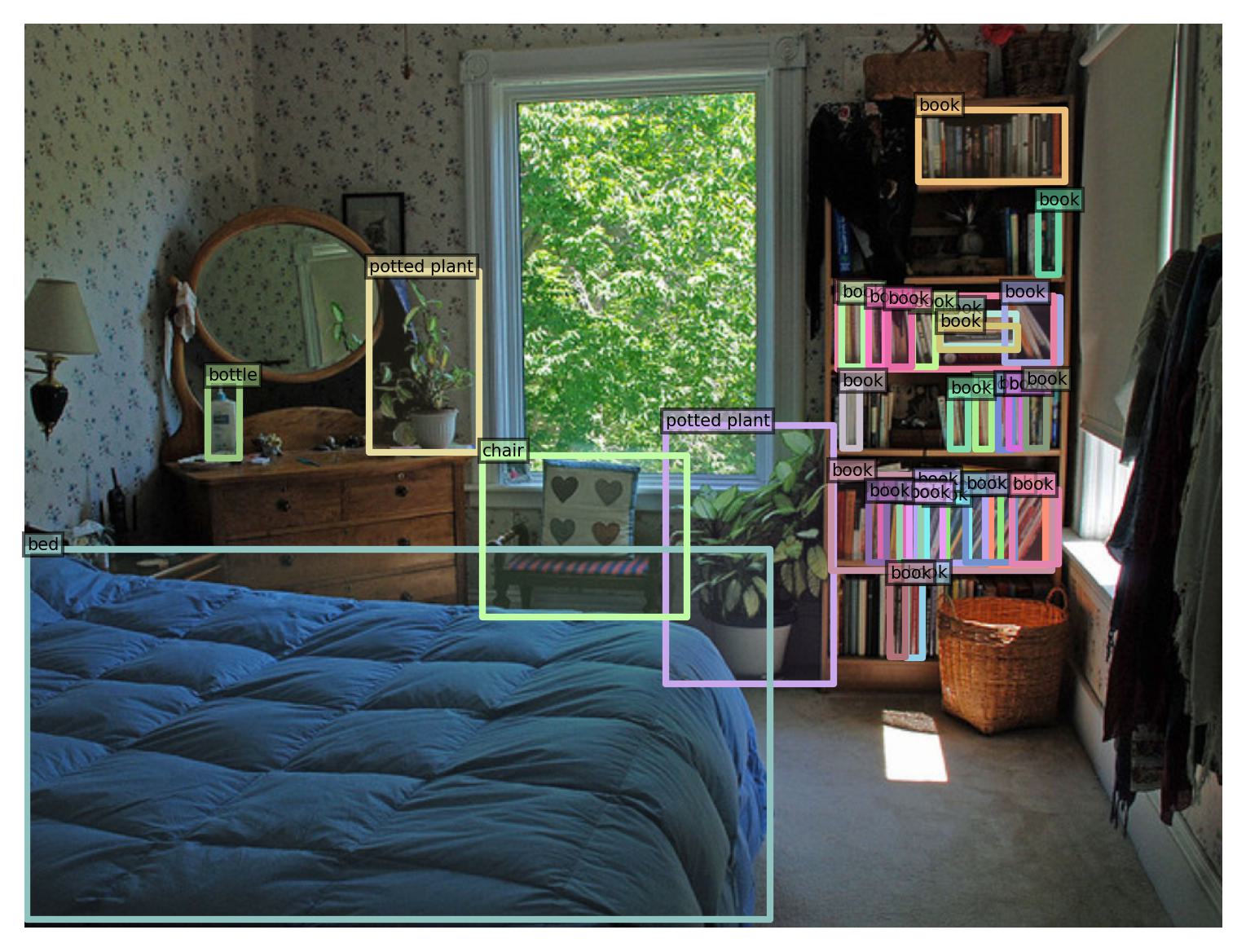}};

\node (im1) [xshift=2*\imw ex, yshift=0*(\imh ex+0.75ex)]{\includegraphics[width=\imw ex, height=\imh ex]{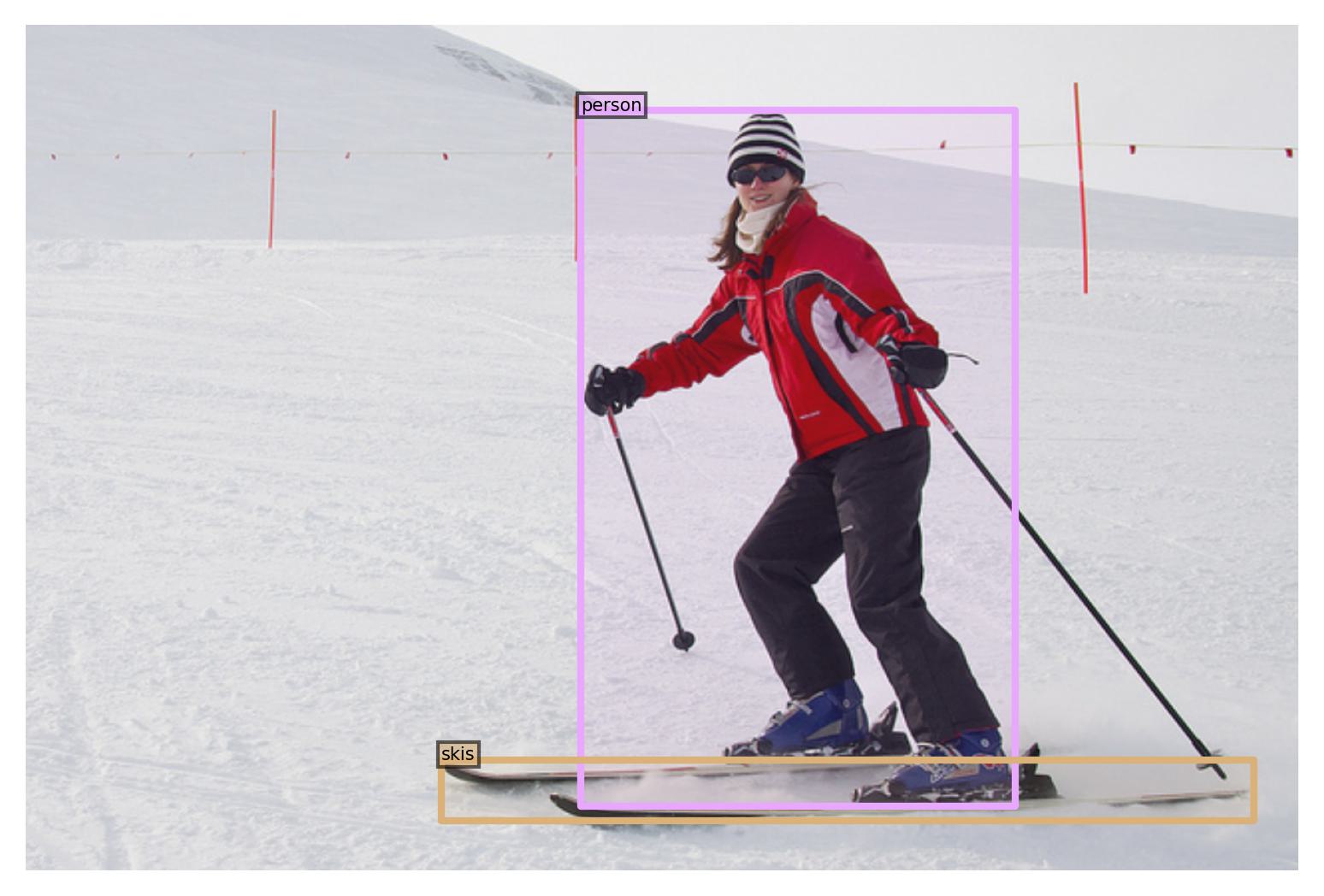}};

\node (im2) [xshift=0 ex, yshift=-1*(\imh ex+0.75ex)]{\includegraphics[width=\imw ex, height=\imh ex]{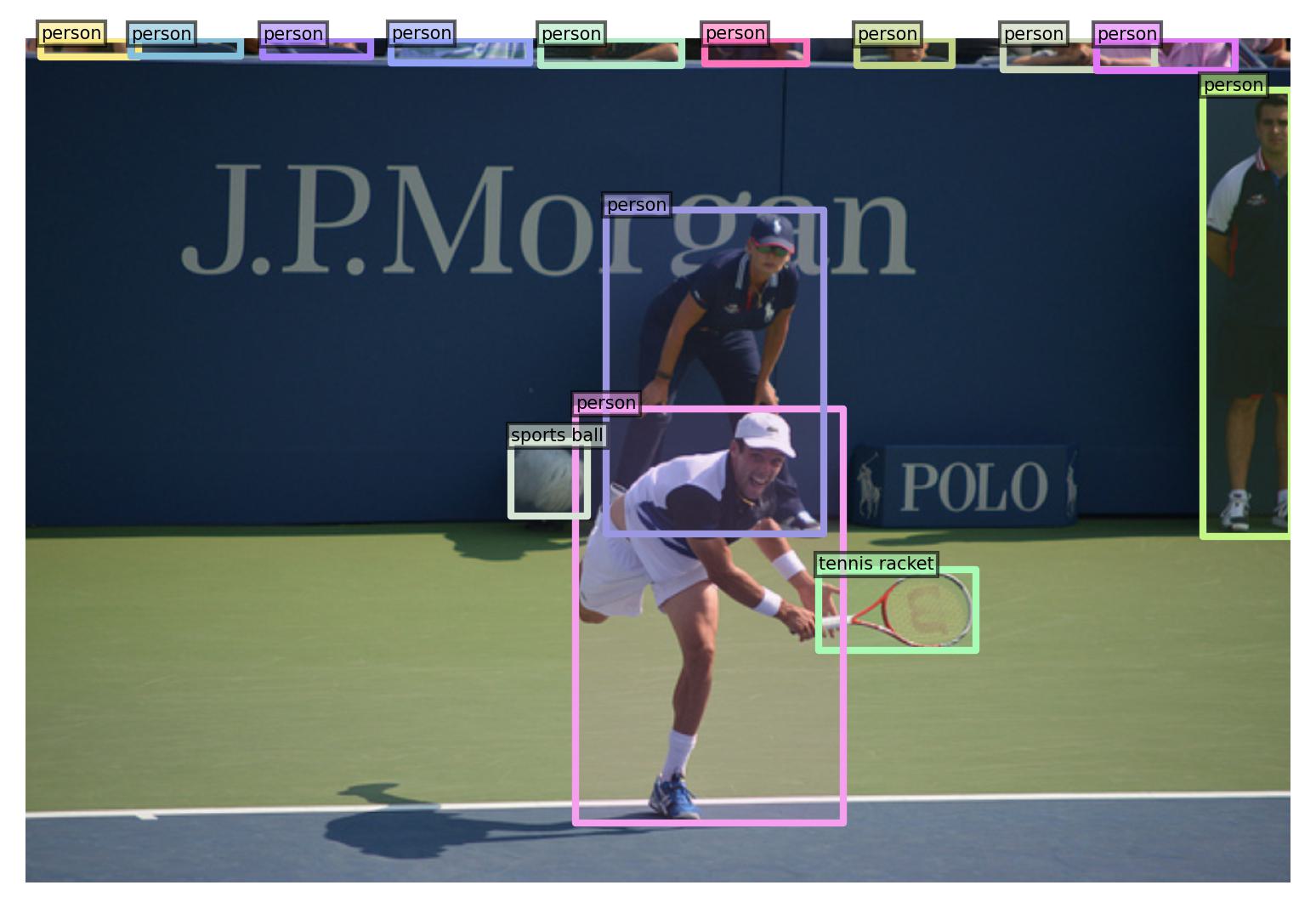}};

\node (im2) [xshift=\imw ex, yshift=-1*(\imh ex+0.75ex)]{\includegraphics[width=\imw ex, height=\imh ex]{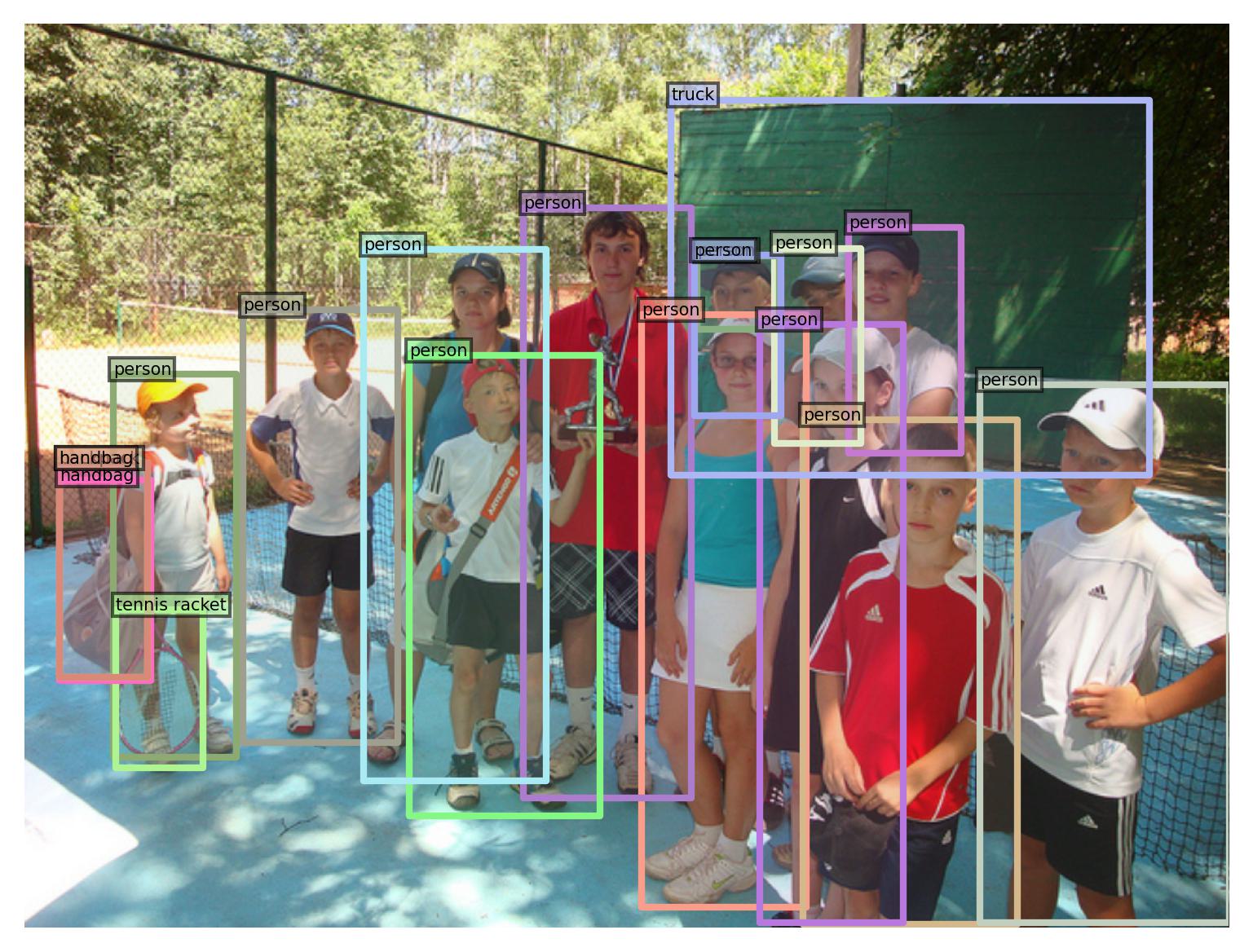}};

\node (im2) [xshift=2*\imw ex, yshift=-1*(\imh ex+0.75ex)]{\includegraphics[width=\imw ex, height=\imh ex]{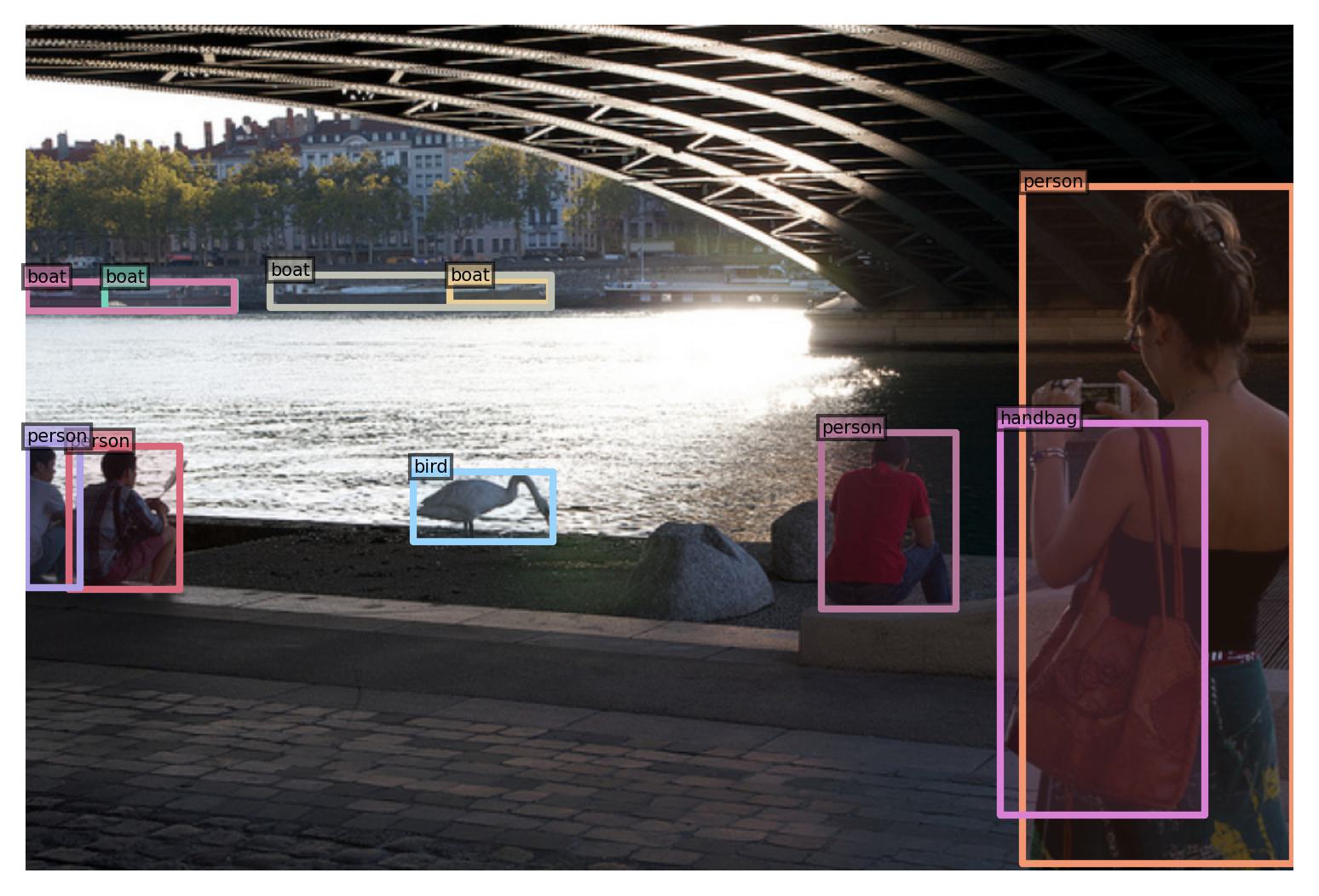}};

\node (im2) [xshift=0 ex, yshift=-2*(\imh ex+0.75ex)]{\includegraphics[width=\imw ex, height=\imh ex]{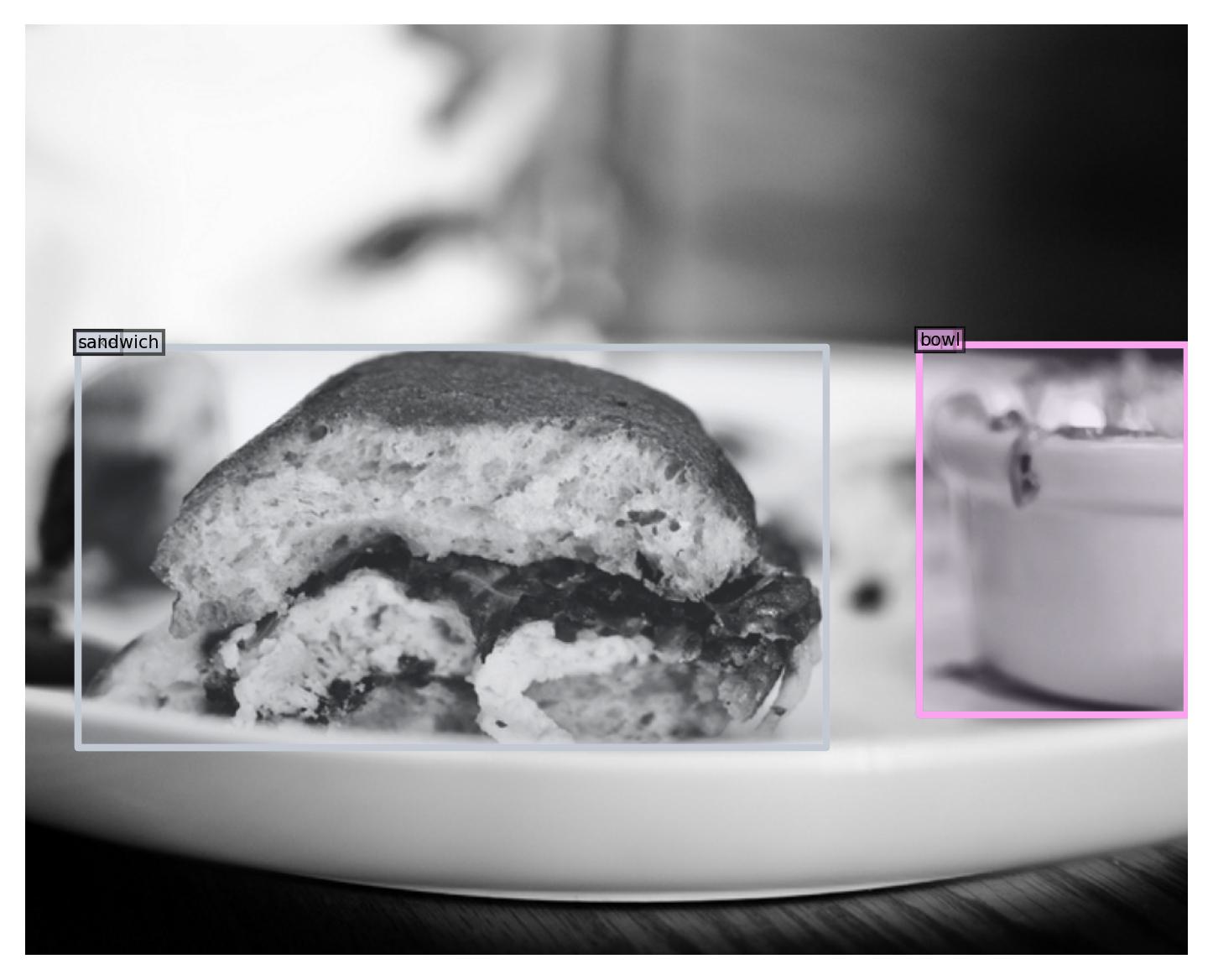}};

\node (im2) [xshift=\imw ex, yshift=-2*(\imh ex+0.75ex)]{\includegraphics[width=\imw ex, height=\imh ex]{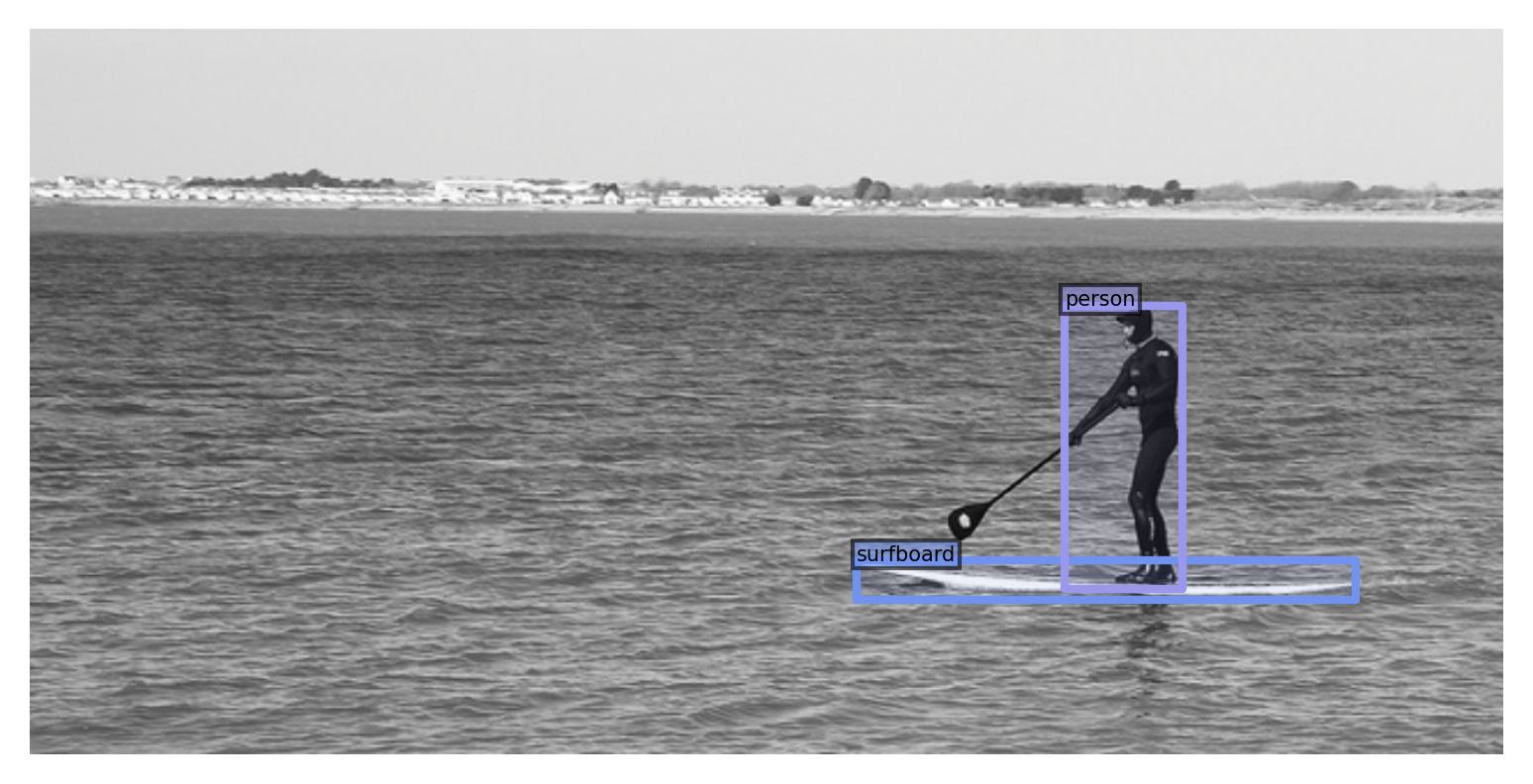}};

\node (im2) [xshift=2*\imw ex, yshift=-2*(\imh ex+0.75ex)]{\includegraphics[width=\imw ex, height=\imh ex]{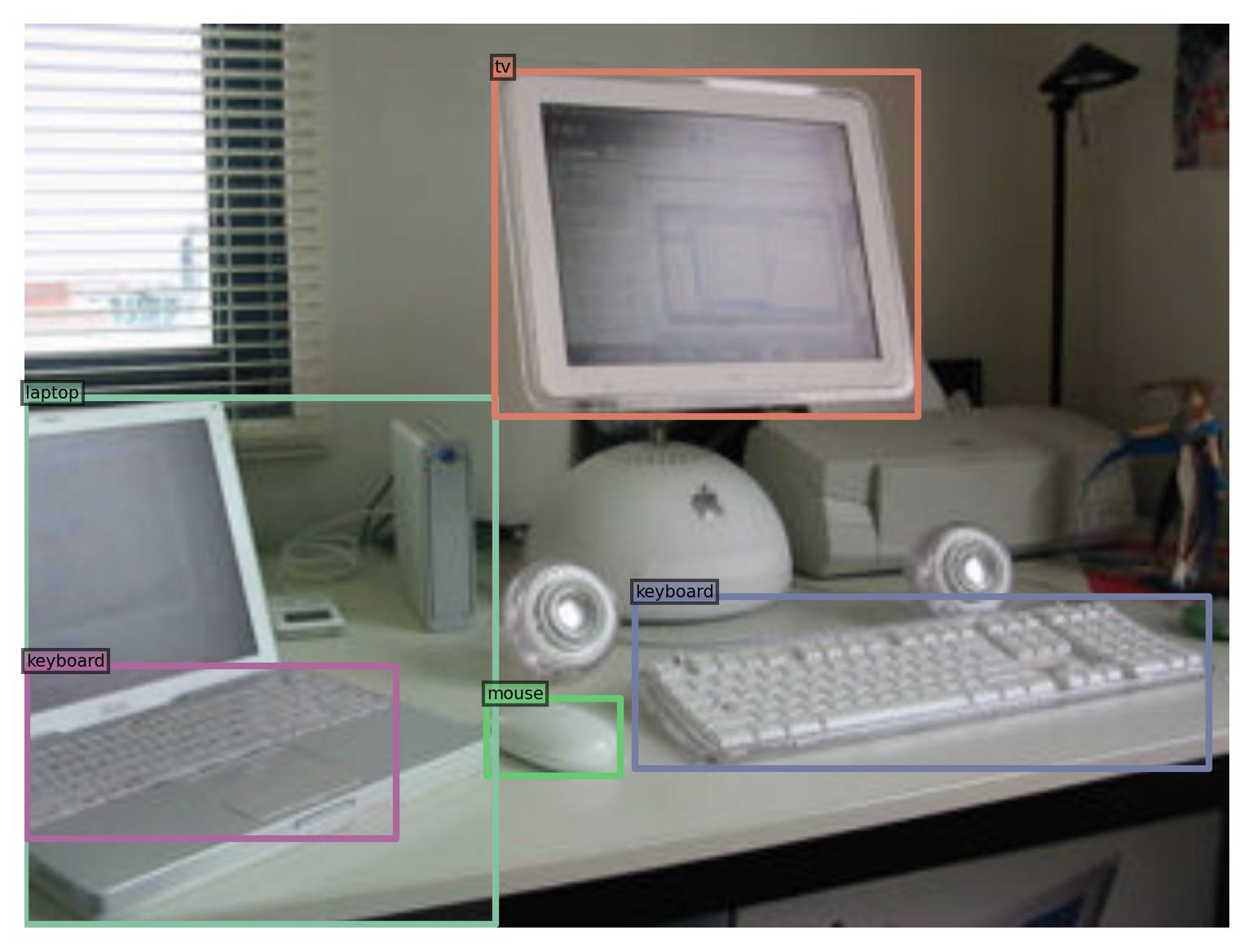}};

\node (im2) [xshift=0 ex, yshift=-3*(\imh ex+0.75ex)]{\includegraphics[width=\imw ex, height=\imh ex]{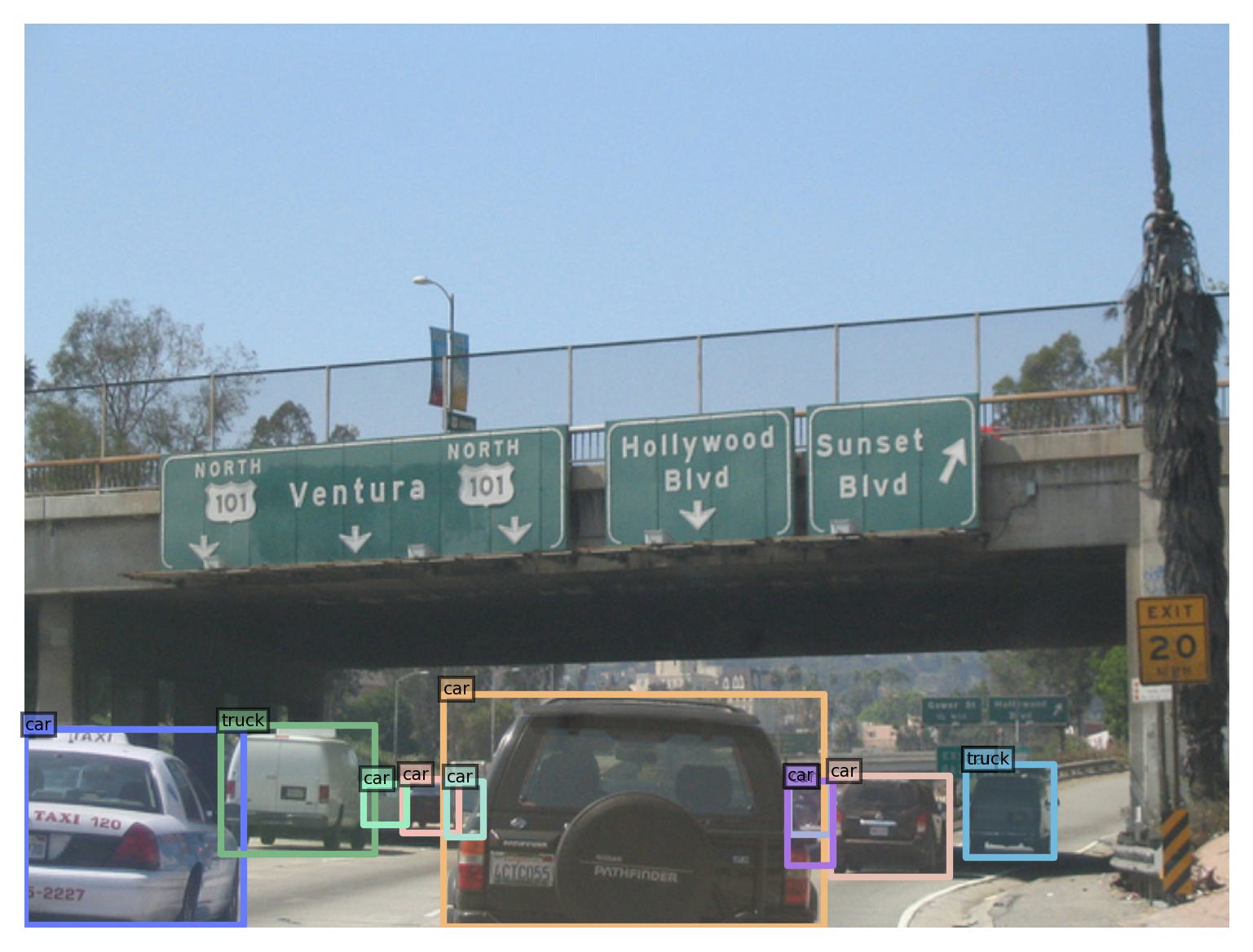}};

\node (im2) [xshift=\imw ex, yshift=-3*(\imh ex+0.75ex)]{\includegraphics[width=\imw ex, height=\imh ex]{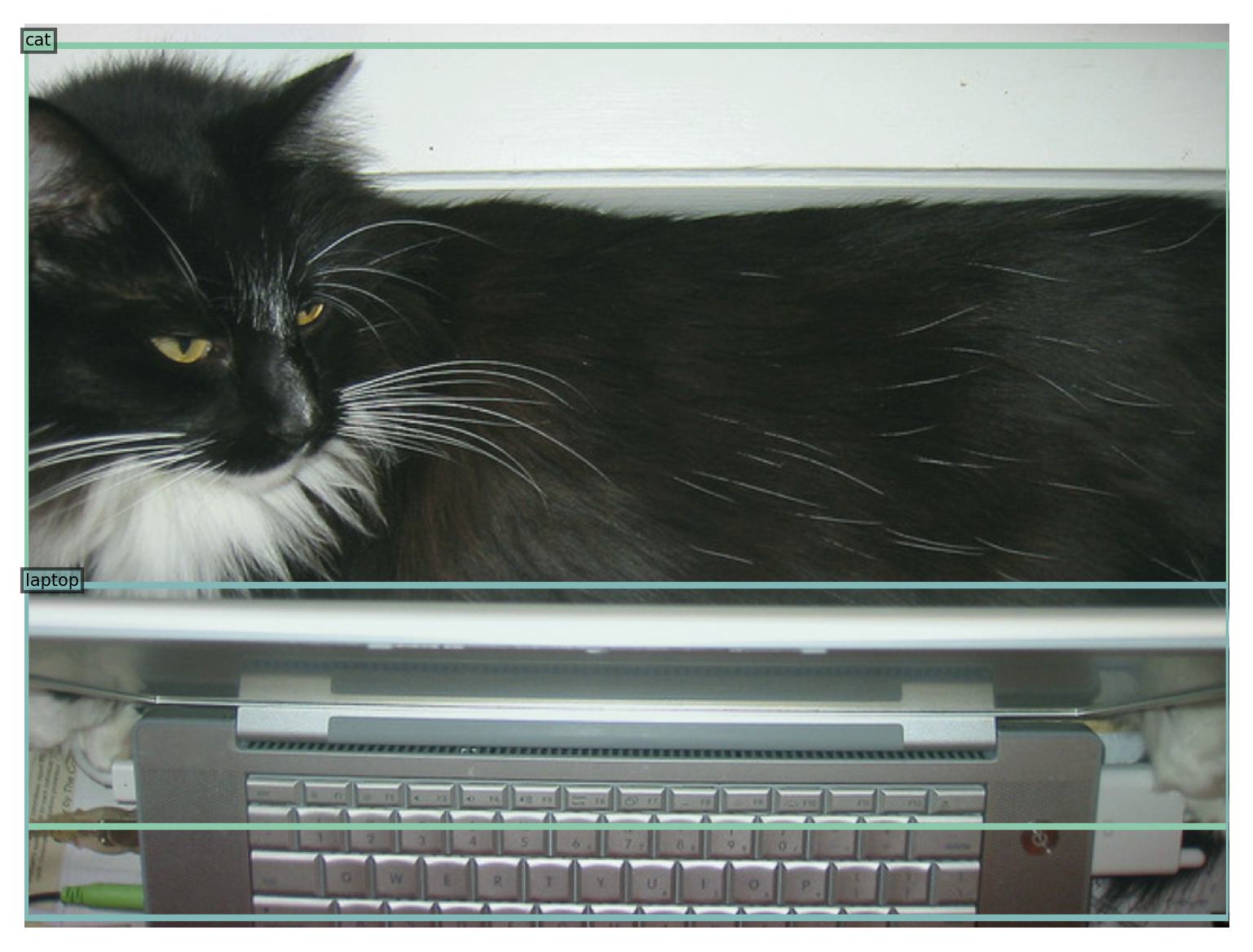}};

\node (im2) [xshift=2*\imw ex, yshift=-3*(\imh ex+0.75ex)]{\includegraphics[width=\imw ex, height=\imh ex]{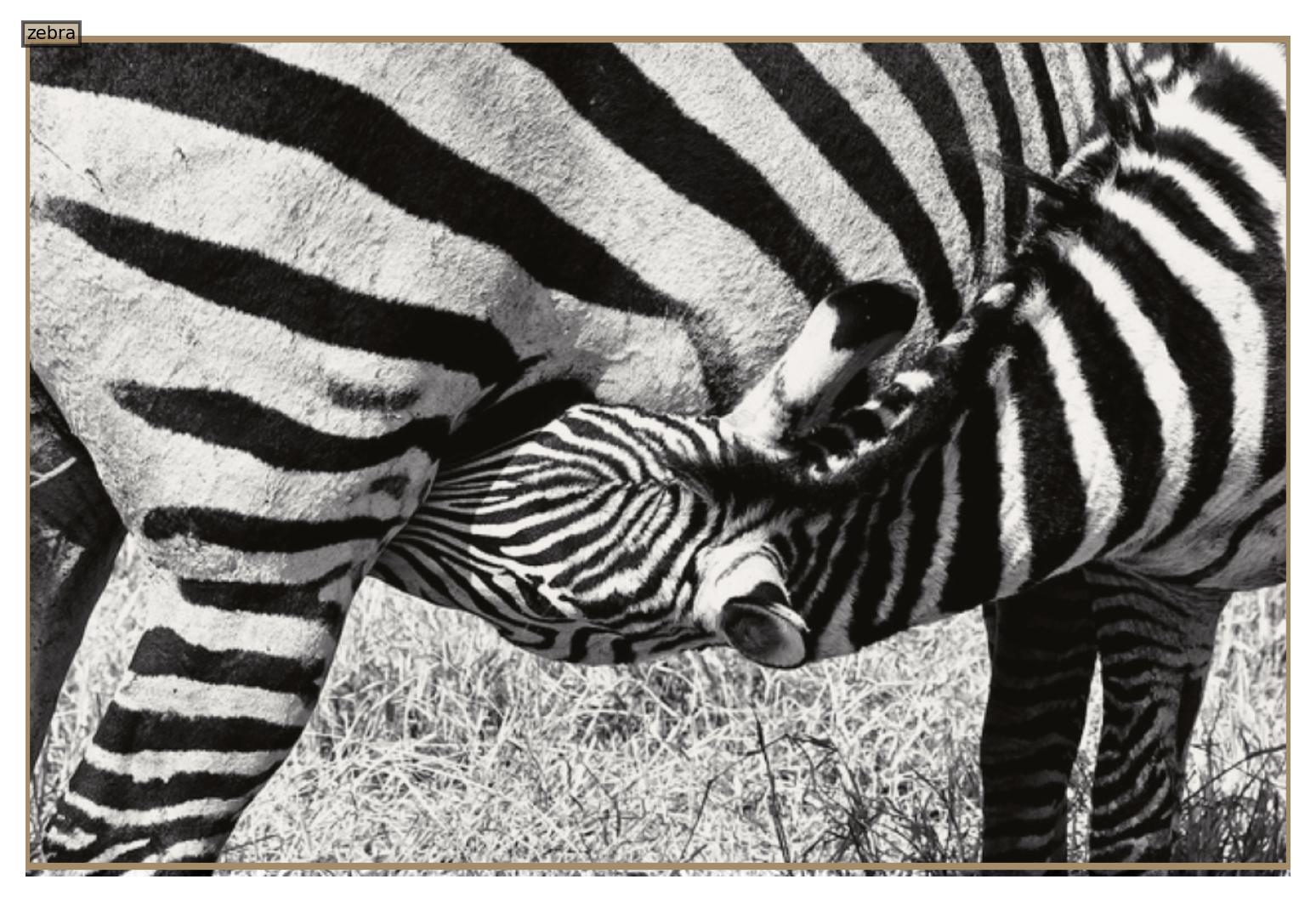}};

\node (im2) [xshift=0 ex, yshift=-4*(\imh ex+0.75ex)]{\includegraphics[width=\imw ex, height=\imh ex]{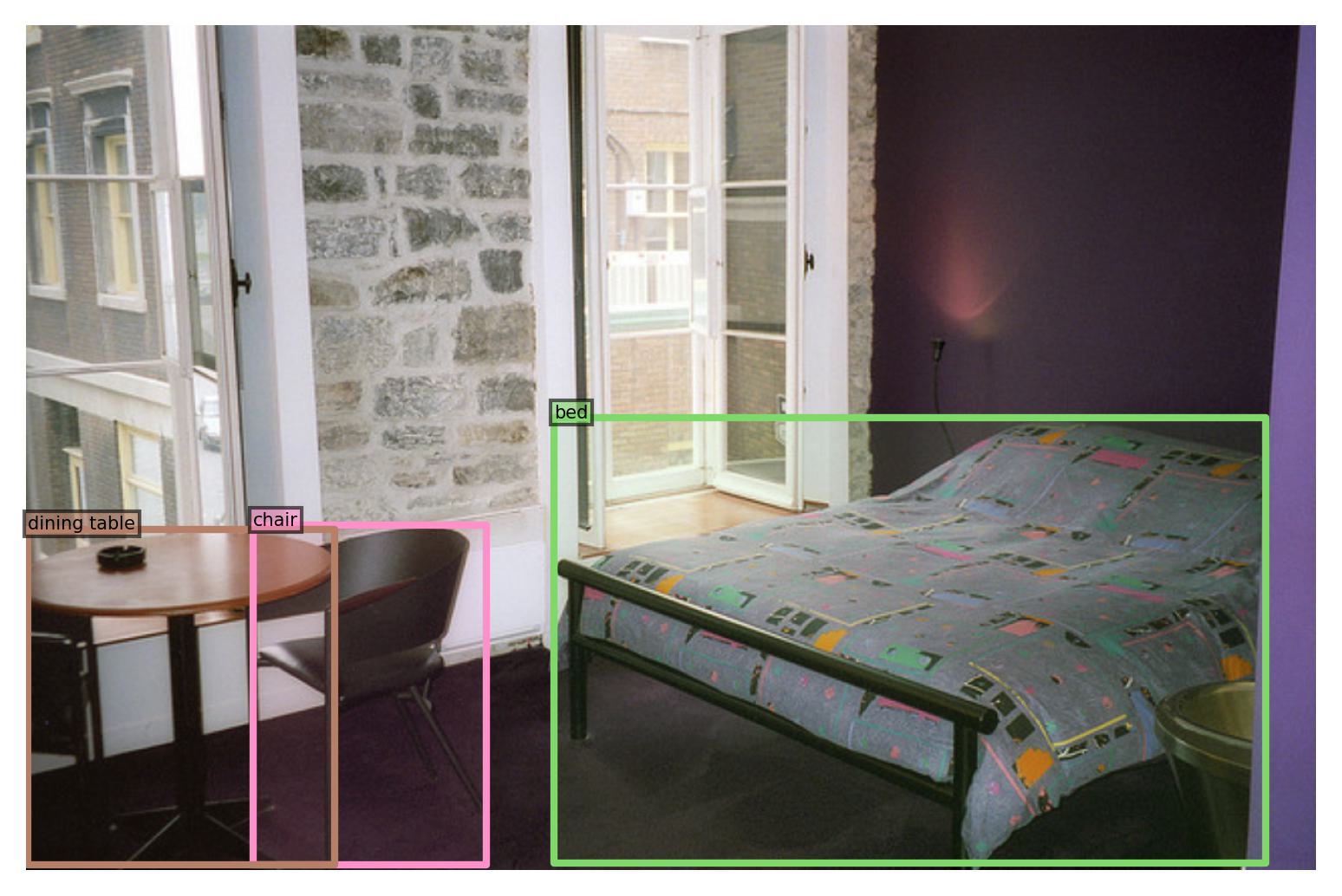}};

\node (im2) [xshift=\imw ex, yshift=-4*(\imh ex+0.75ex)]{\includegraphics[width=\imw ex, height=\imh ex]{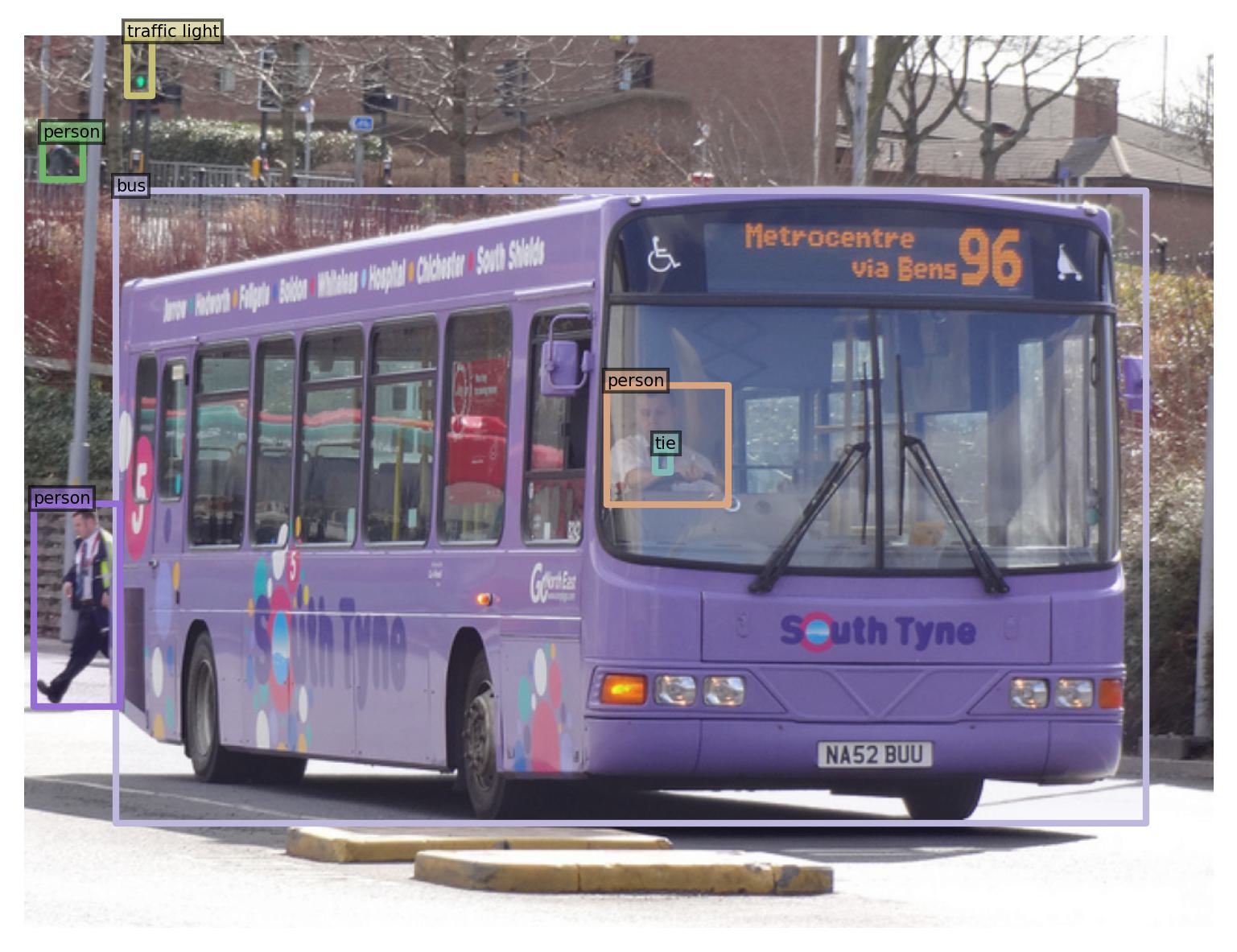}};

\node (im2) [xshift=2*\imw ex, yshift=-4*(\imh ex+0.75ex)]{\includegraphics[width=\imw ex, height=\imh ex]{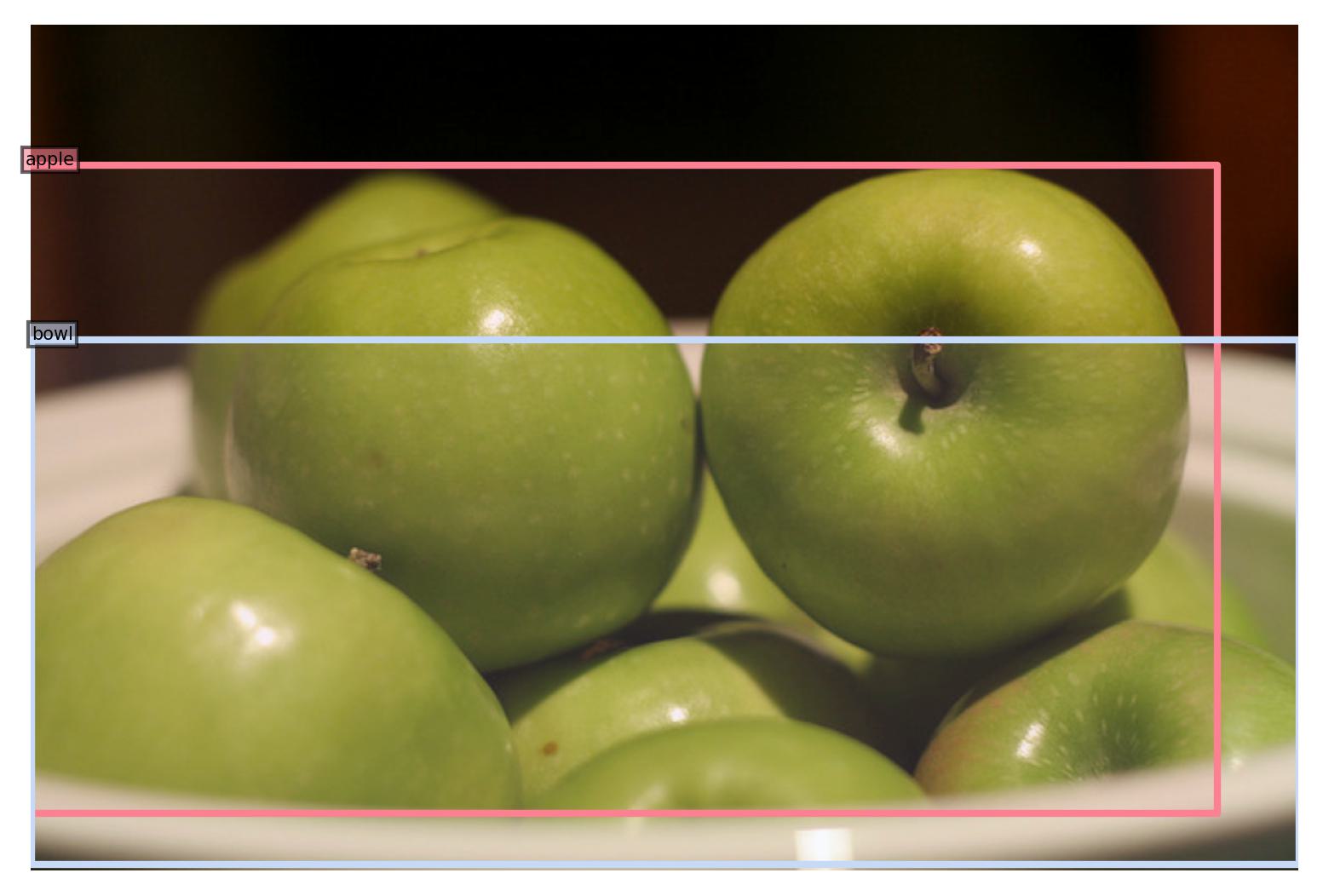}};

\node (im2) [xshift=0 ex, yshift=-5*(\imh ex+0.75ex)]{\includegraphics[width=\imw ex, height=\imh ex]{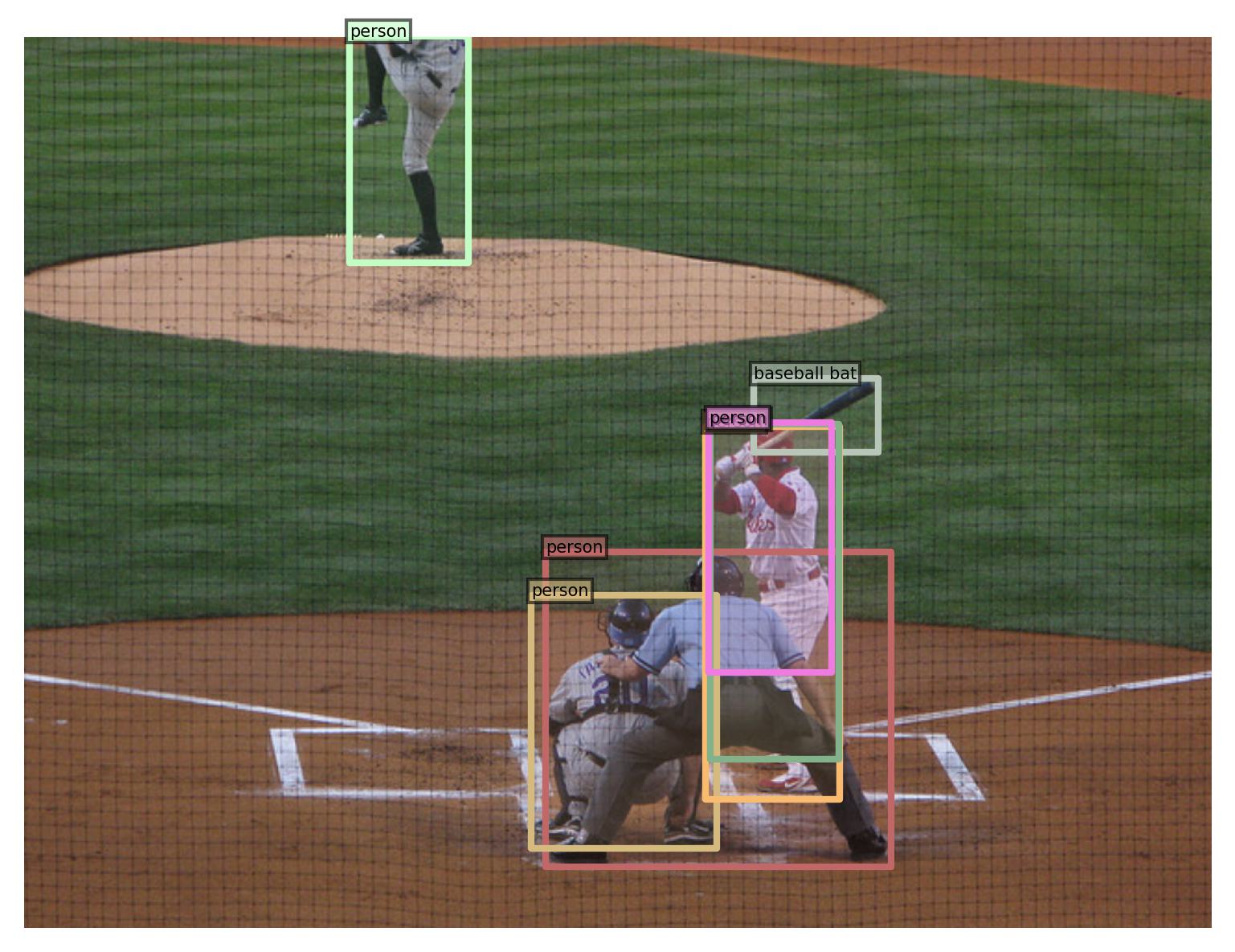}};

\node (im2) [xshift=\imw ex, yshift=-5*(\imh ex+0.75ex)]{\includegraphics[width=\imw ex, height=\imh ex]{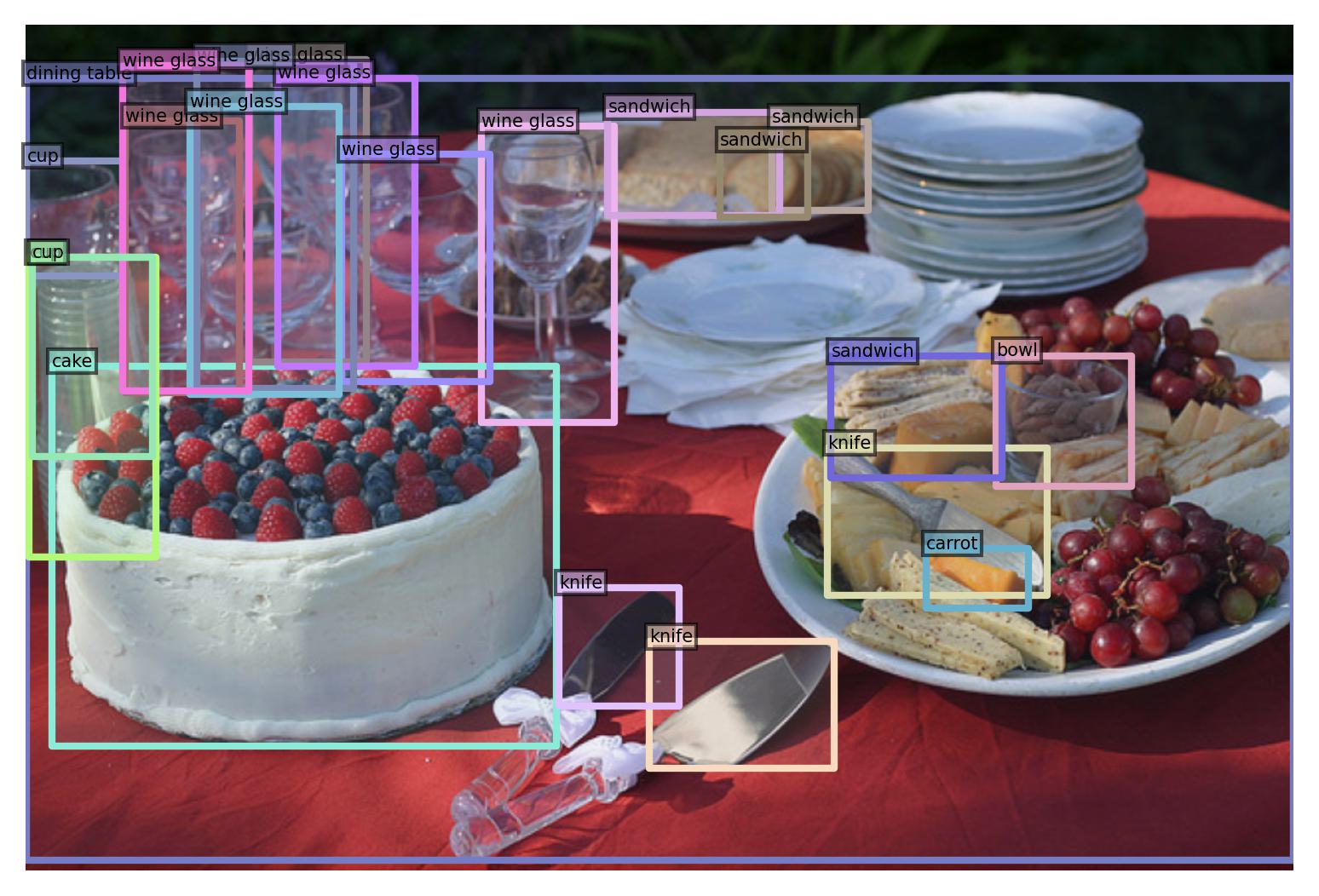}};

\node (im2) [xshift=2*\imw ex, yshift=-5*(\imh ex+0.75ex)]{\includegraphics[width=\imw ex, height=\imh ex]{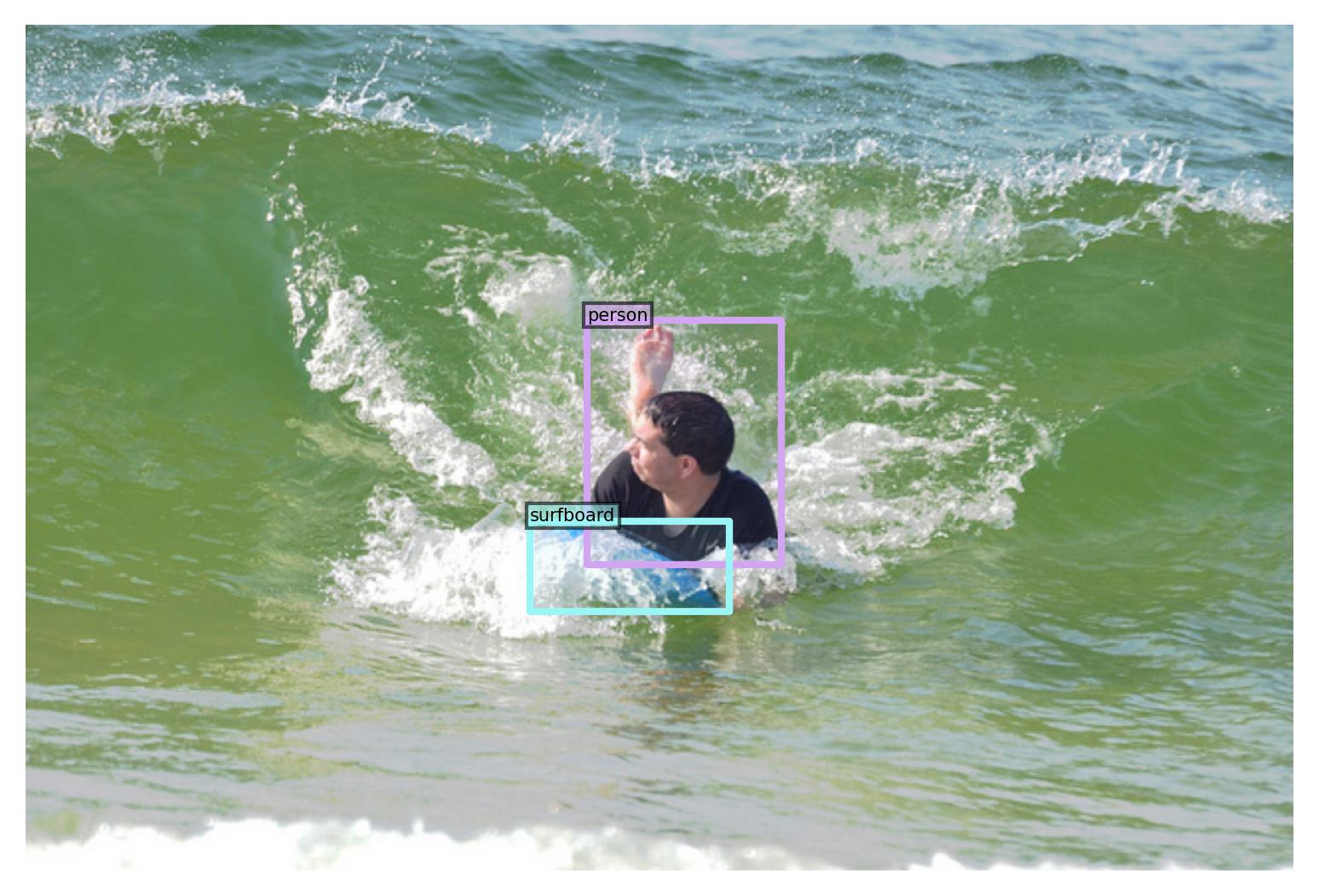}};

\end{tikzpicture}
\vspace{-1.0ex}
%
%\caption{Detection results from MS-COCO validations set.}
%
\label{fig:det}

\vspace{-3ex}
\end{figure}
% \end{wrapfigure}
%

\newpage

\begin{figure}[!h]
% \begin{wrapfigure}{r}{0.40\textwidth}

\centering

\begin{tikzpicture}

\FPeval{\imw}{30}
\FPeval{\imh}{30}

\node (im2) [xshift=0 ex, yshift=0*(\imh ex+0.75ex)]{\includegraphics[width=\imw ex, height=\imh ex]{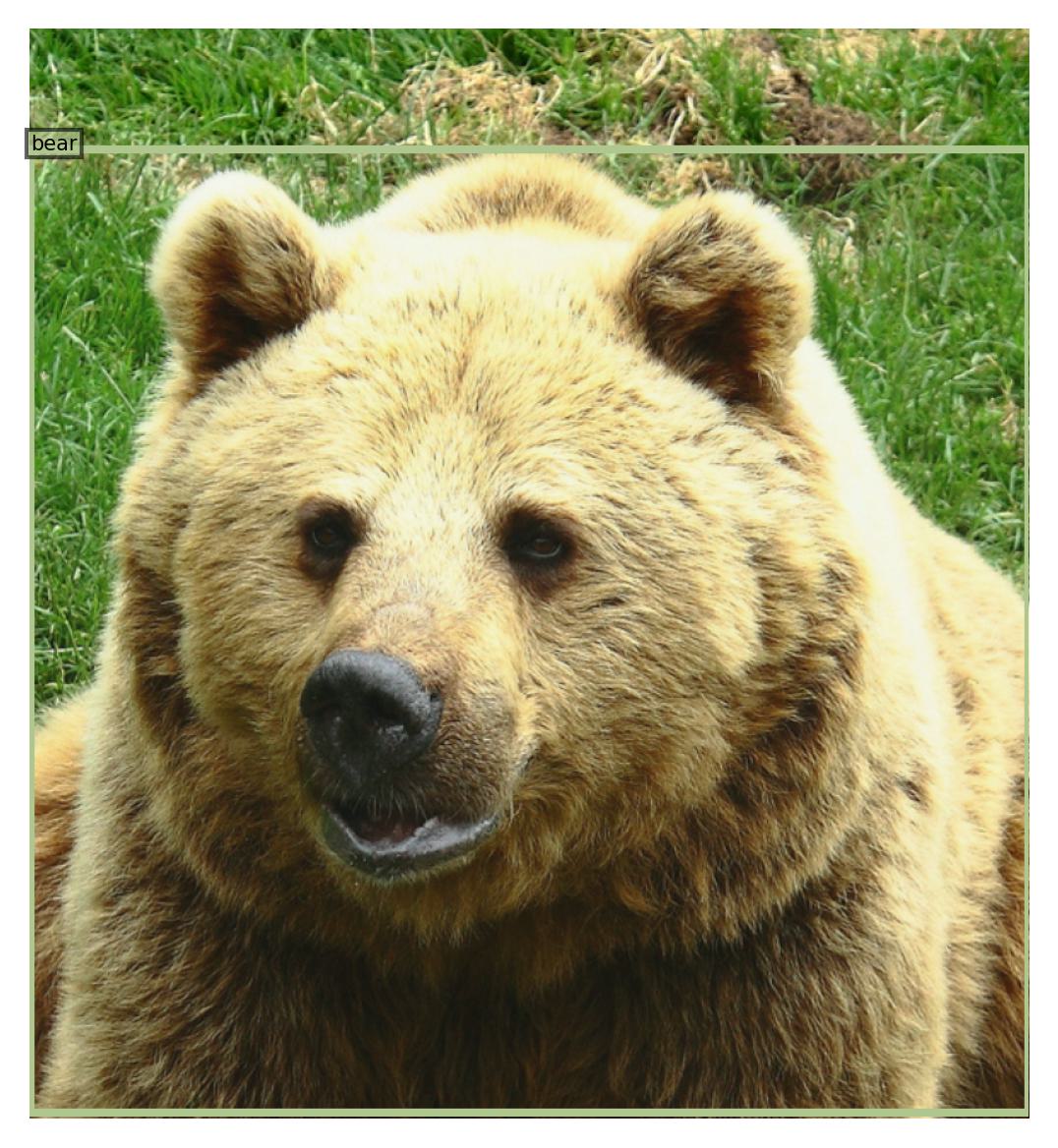}};

\node (im1) [xshift=1*\imw ex, yshift=0*(\imh ex+0.75ex)]{\includegraphics[width=\imw ex, height=\imh ex]{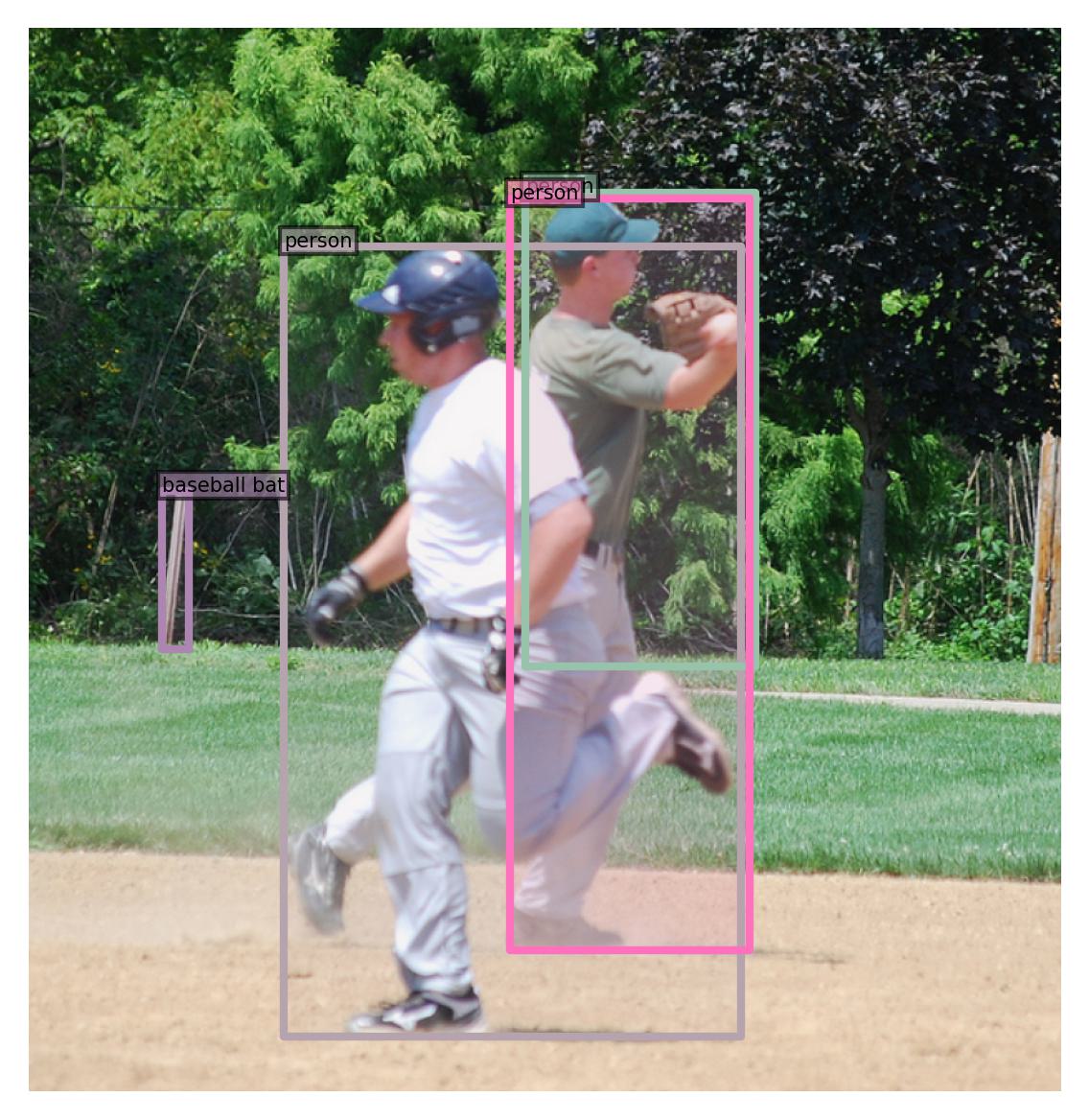}};

\node (im1) [xshift=2*\imw ex, yshift=0*(\imh ex+0.75ex)]{\includegraphics[width=\imw ex, height=\imh ex]{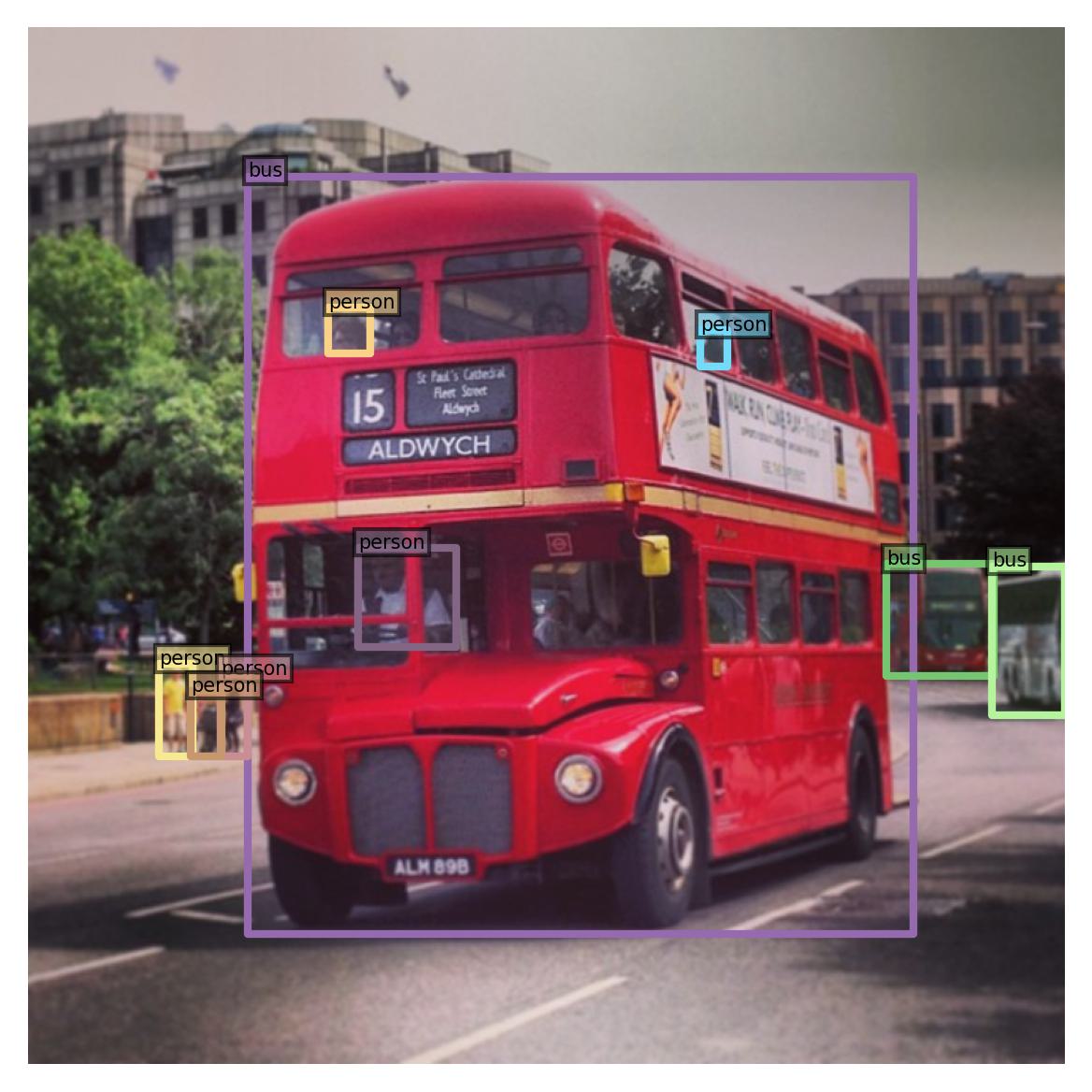}};

\FPeval{\imw}{30}
\FPeval{\imh}{40}

\node (im1) [xshift=0ex, yshift=-1*(\imh ex-4.8ex)]{\includegraphics[width=\imw ex, height=\imh ex]{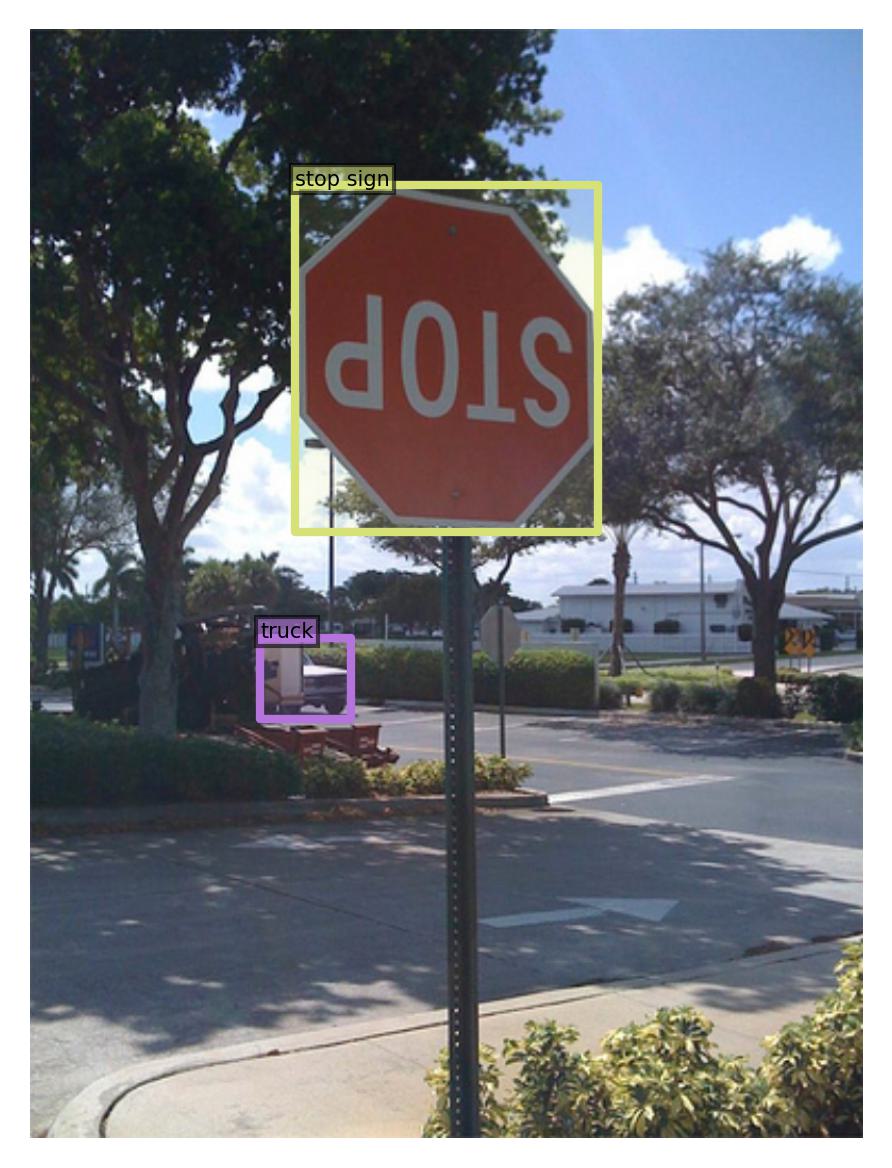}};

\node (im2) [xshift=\imw ex, yshift=-1*(\imh ex-4.8ex)]{\includegraphics[width=\imw ex, height=\imh ex]{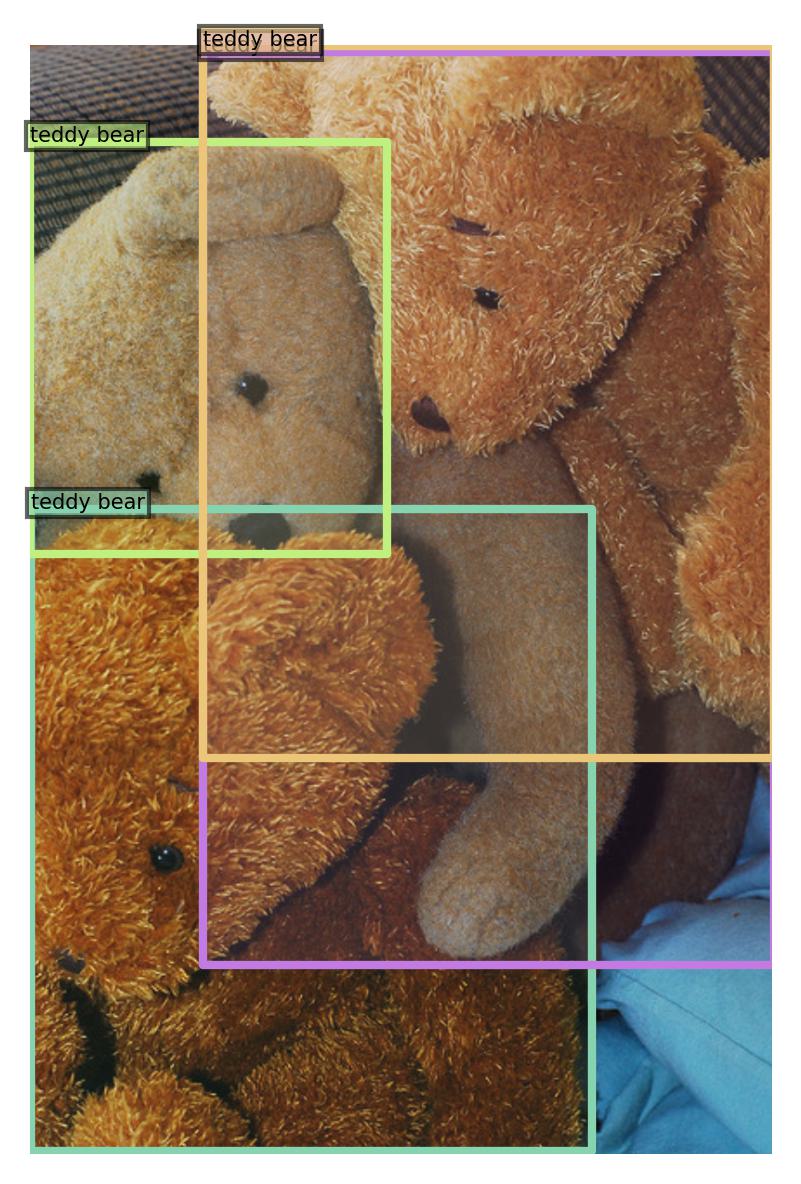}};

\node (im1) [xshift=2*\imw ex, yshift=-1*(\imh ex-4.8ex)]{\includegraphics[width=\imw ex, height=\imh ex]{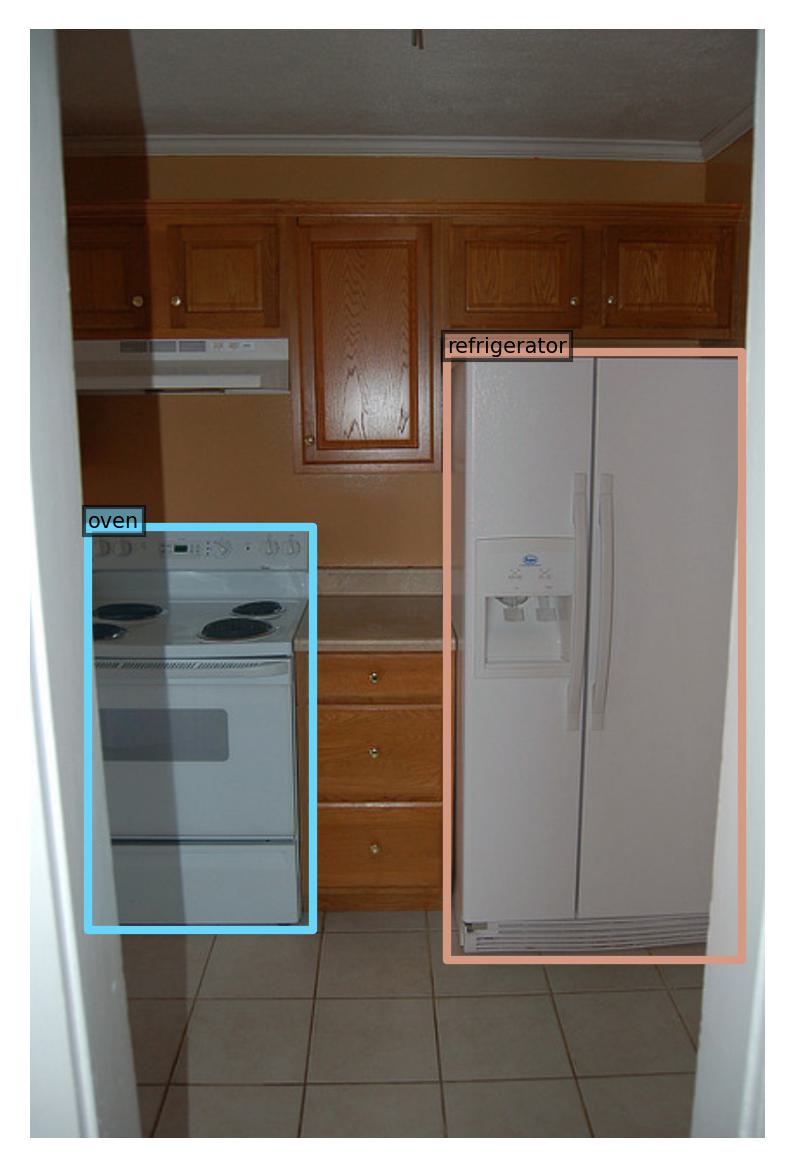}};

\node (im1) [xshift=0ex, yshift=-2*(\imh ex-2.75ex)]{\includegraphics[width=\imw ex, height=\imh ex]{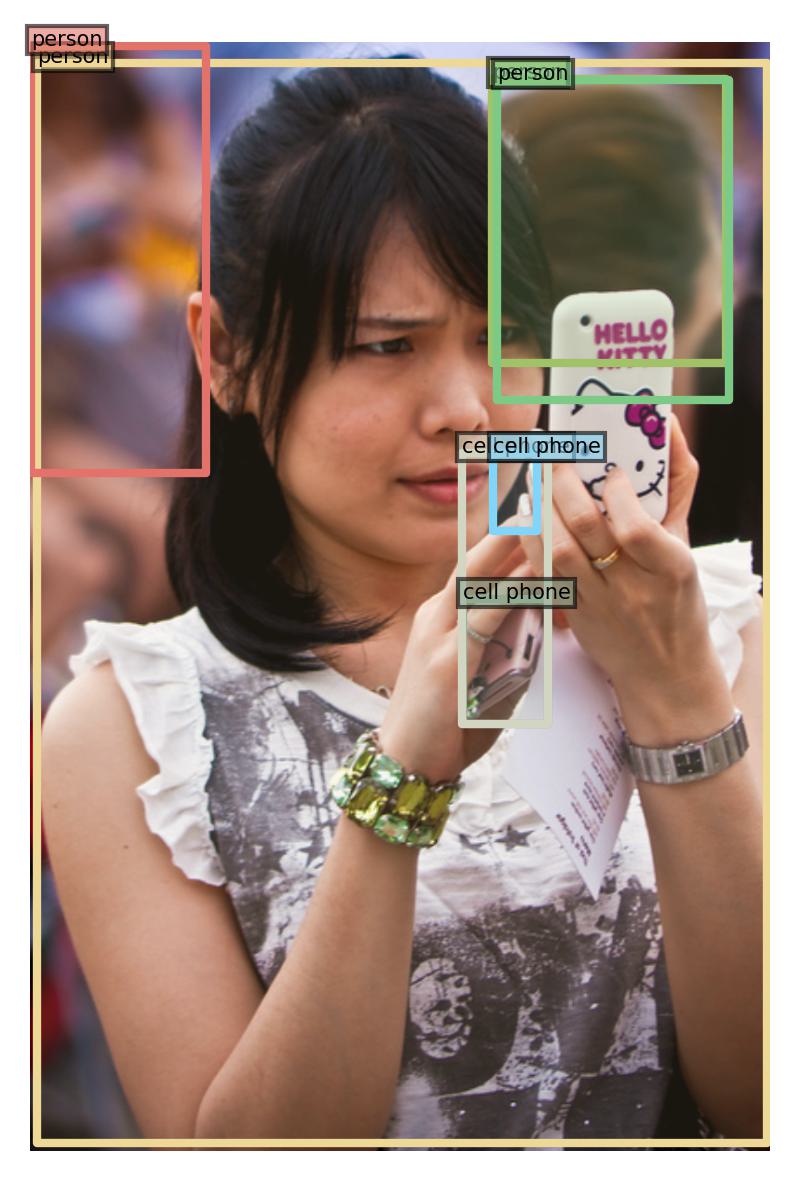}};

\node (im2) [xshift=\imw ex, yshift=-2*(\imh ex-2.75ex)]{\includegraphics[width=\imw ex, height=\imh ex]{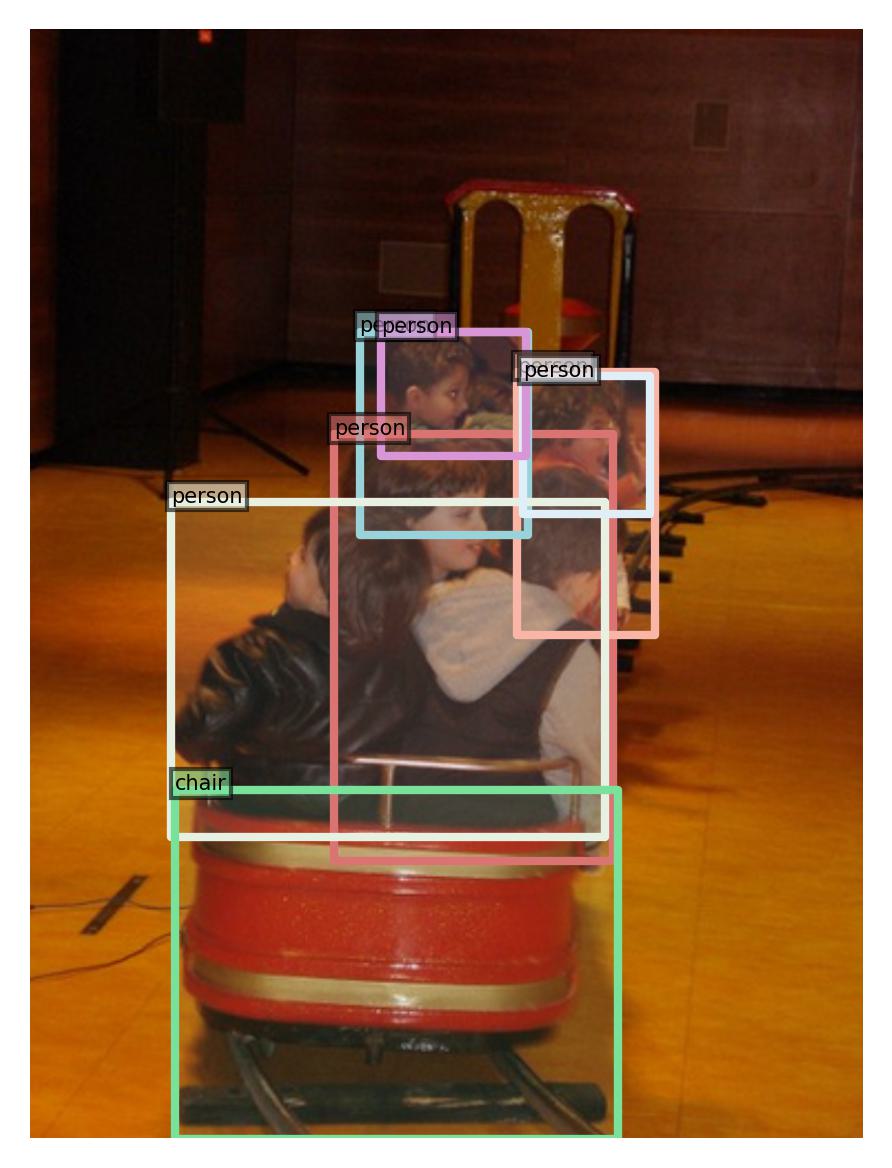}};

\node (im1) [xshift=2*\imw ex, yshift=-2*(\imh ex-2.75ex)]{\includegraphics[width=\imw ex, height=\imh ex]{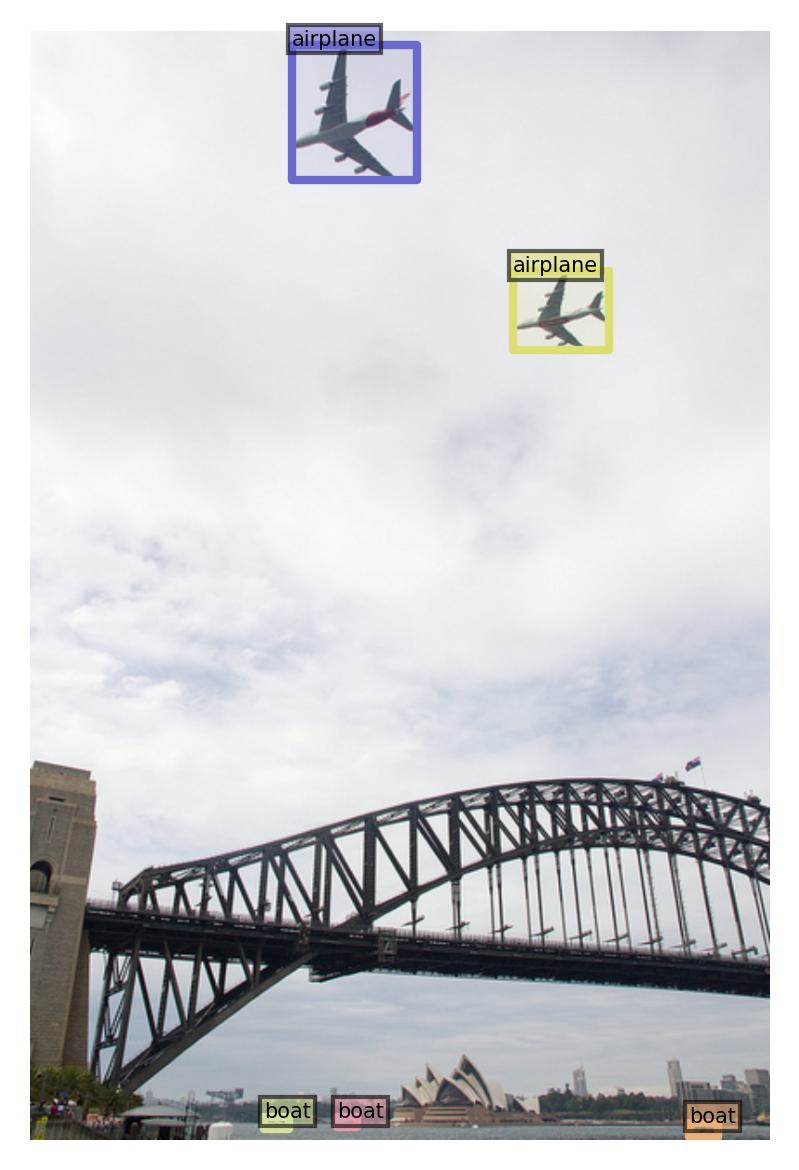}};

%
% \node (im1) [xshift=0ex, yshift=-3*(\imh ex+0.75ex)]{\includegraphics[width=\imw ex, height=\imh ex]{detections/28.jpg}};

\end{tikzpicture}
\vspace{-1.0ex}
%
%\caption{Detection results from MS-COCO validations set.}
%
\label{fig:det}

\vspace{-3ex}
\end{figure}
% \end{wrapfigure}
%

\end{document}